\documentclass[journal=jpccck,manuscript=article,maxauthors=99,etalmode=truncate]{achemso}

\usepackage{macros}

\usepackage{header}

\usepackage[version=4]{mhchem}

\begin{document}

\begin{abstract}
  Graph neural networks (GNNs) have been shown to be astonishingly capable models for molecular property prediction, particularly as surrogates for expensive density functional theory calculations of relaxed energy for novel material discovery.
  However, one limitation of GNNs in this context is the lack of useful uncertainty prediction methods, as this is critical to the material discovery pipeline.
  In this work, we show that uncertainty quantification for relaxed energy calculations is more complex than uncertainty quantification for other kinds of molecular property prediction, due to the effect that structure optimizations have on the error distribution.
  We propose that distribution-free techniques are more useful tools for assessing calibration, recalibrating, and developing uncertainty prediction methods for GNNs performing relaxed energy calculations.
  We also develop a relaxed energy task for evaluating uncertainty methods for equivariant GNNs, based on distribution-free recalibration and using the Open Catalyst Project dataset.
  We benchmark a set of popular uncertainty prediction methods on this task, and show that conformal prediction methods, with our novel latent space distance improvements, are the most well-calibrated and efficient approach for relaxed energy calculations.
  Finally, we demonstrate that our latent space distance method produces results which align with our expectations on a atom clustering example, and on specific equation of state and adsorbate coverage examples from outside the training dataset.

\end{abstract}

\section{Introduction}

To keep up with growing energy demands, it is necessary to search for novel catalyst materials to enable more efficient storage of renewable sources of energy \cite{zitnick2020introduction, ocp, norskovbook, Dumesic}. 
Computational material discovery is crucial to this process, as it enables less expensive screening of an enormous space of possible catalyst materials than physical experiments.
Faster and more accurate computational material discovery methods will be required to meet our society's renewable energy needs in the face of a rapidly changing climate.

Graph neural networks are state of the art in accelerating computational material discovery pipelines with machine learning potentials. 
Machine learning potentials work as surrogate models trained to approximate computationally expensive \gls{dft} calculations of energy and forces on atomistic structures.
This task is referred to as \gls{s2ef}.
These energy and force calculations are used to iteratively perform geometric optimizations of atomic positions (referred to in this work as "relaxations"), to minimize their energy.
These relaxed structure and relaxed energy calculations are what enable high-throughput predictions of catalyst performance in the real world.
For a given catalyst-adsorbate system, the global minimum relaxed energy (adsorption energy) directly correlates with the reactivity and selectivity of reaction pathways on that catalyst surface \cite{tran2018active, zhong2020accelerated, liu2017understanding, norskov2005trends, wan2022machine, ulissi2017address}.

In recent years, \gls{gnn}s have made tremendous strides in replacing \gls{dft} codes with an inexpensive, accurate alternative \cite{gemnetoc, scn, escn, equiformerv2}.
Thanks to methods like AdsorbML, \gls{gnn}s can speed up adsorption energy calculations alone at the cost of accuracy, or in tandem with \gls{dft} at the cost of speed \cite{adsorbml}. 
The recent OCP Demo (\url{https://open-catalyst.metademolab.com}) is a publicly available tool where \gls{gnn}s are used to calculate these adsorption energies without any expensive \gls{dft} calculations.
However, a major limitation of current \gls{gnn}s is their lack of uncertainty estimates for relaxed energy predictions. 
Ideally, users of these methods would know when it is safe to trust the \gls{gnn} predictions, and when additional \gls{dft} calculations are warranted.
In this work, we specifically examine methods of \gls{uq} estimates for \gls{gnn} predictions on this \gls{rs2re} task.

\section{Background}

\subsection{AdsorbML}

AdsorbML \cite{adsorbml} is a method to calculate adsorption energy using machine learning potentials.
In order to find the global minimum relaxed energy for a specified surface and adsorbate, this method places the adsorbate in many different starting configurations, relaxes each configuration, and returns the minimum of all the relaxed energies.
Traditionally, this would be done using an \textit{ab initio} method such as \gls{dft}, but \gls{dft} is very costly and this approach is infeasibly expensive.
AdsorbML uses \gls{gnn}s as a surrogate for \gls{dft} to perform the relaxations instead, requiring only a solitary \gls{dft} single point calculation for the relaxed structure to verify the relaxed energy.
The required number of expensive \gls{dft} calculations is further reduced by using \gls{gnn}s to filter out all but the few most promising candidates.
This method provides an adjustable spectrum of trade-offs between accuracy and efficiency, with one balanced option finding an equivalent or better adsorption energy 87.36\% of the time while reducing \gls{dft} compute by more than a factor of 2000.
Ideally, no \gls{dft} would be required, but even state of the art \gls{gnn}s are unreliable energy predictors when out of domain, and using them alone drops the success rate to 56\%.
Without \gls{dft}, such as in the OCP demo, we need uncertainty metrics, so users know when to trust the results of these models.

\subsection{Graph Neural Networks}

This work focuses on quantifying uncertainty prediction methods for EquiformerV2 \cite{equiformerv2}, a \gls{gnn} model architecture for molecular property prediction. We choose EquiformerV2 because it is the current state of the art in molecular property prediction for catalyst materials, according to the \gls{ocp} leaderboard\cite{ocp}. We also compare it to Gemnet-OC \cite{gemnetoc}, another high performing \gls{gnn} on the leaderboard. Both of these models are used in the \gls{ocp} Demo, to run the AdsorbML algorithm and predict minimum relaxed energies without the use of expensive \gls{dft} calculations\cite{adsorbml}.

\subsection{Uncertainty Quantification}

Many prior studies have examined the application of \gls{uq} techniques to machine learning potentials and molecular property prediction.
Most \gls{uq} validation metrics seek to measure some description of the calibration of an uncertainty prediction method.
The most popular \gls{uq} validation metrics are miscalibration area, Spearman's rank correlation coefficient, and the negative log likelihood of the errors given the uncertainties\cite{conformal_prediction, conformal_kulik,uq_tran_benchmark,uq_per_node_gp,uq_mve_deep_ensembles,uq_benchmark_ensembles_bad,uq_benchmark_ensembles_good}.
These metrics each face drawbacks, especially when used for overall validation of uncertainty estimates. 
Specifically, the miscalibration area and \gls{nll} rely on an assumption of normally distributed errors, which adds fragility to the metrics \cite{pernot}. 
Additionally, the \gls{nll} and \spearman validation metrics are not well suited to the absolute validation of uncertainty estimates, which limits them to comparing models and does not allow for the overall validation of the uncertainty estimates \cite{rasmussen_duan_kulik_jensen_2023}.
Notably these metrics each target different properties of uncertainty estimates, and are not consistently in agreement about which uncertainty estimator performs best, even within a single study \cite{uq_coley_benchmark}.

Calibration is the primary \gls{uq} validation metric for making direct comparisons between uncertainty prediction methods.
Prior work by Pernot, and Levi et al. show distribution free methods of measuring local and global miscalibration: the CI(Var(Z)) test and the error-based calibration plot \cite{pernot, levi2022_error_based_cal}.
These approaches can be more effective in describing the performance of predicted uncertainties of surrogate models when the expected error distribution cannot be assumed to be Gaussian.
Rasmussen et al. finds that the error-based calibration plot generally serves as the best overall method of validating of uncertainty estimates for machine learning potentials for chemistry \cite{rasmussen_duan_kulik_jensen_2023}.
Error-based calibration measures can also be used to recalibrate uncertainty predictions, allowing a variety of uncertainty quantification methods to be recalibrated and compared.

\section{Methods}

\subsection{Uncertainty Quantification Methods for GNNs}
\label{sec:uncertainty_methods}

\begin{figure}
\centering
\includegraphics[width=0.99\textwidth]{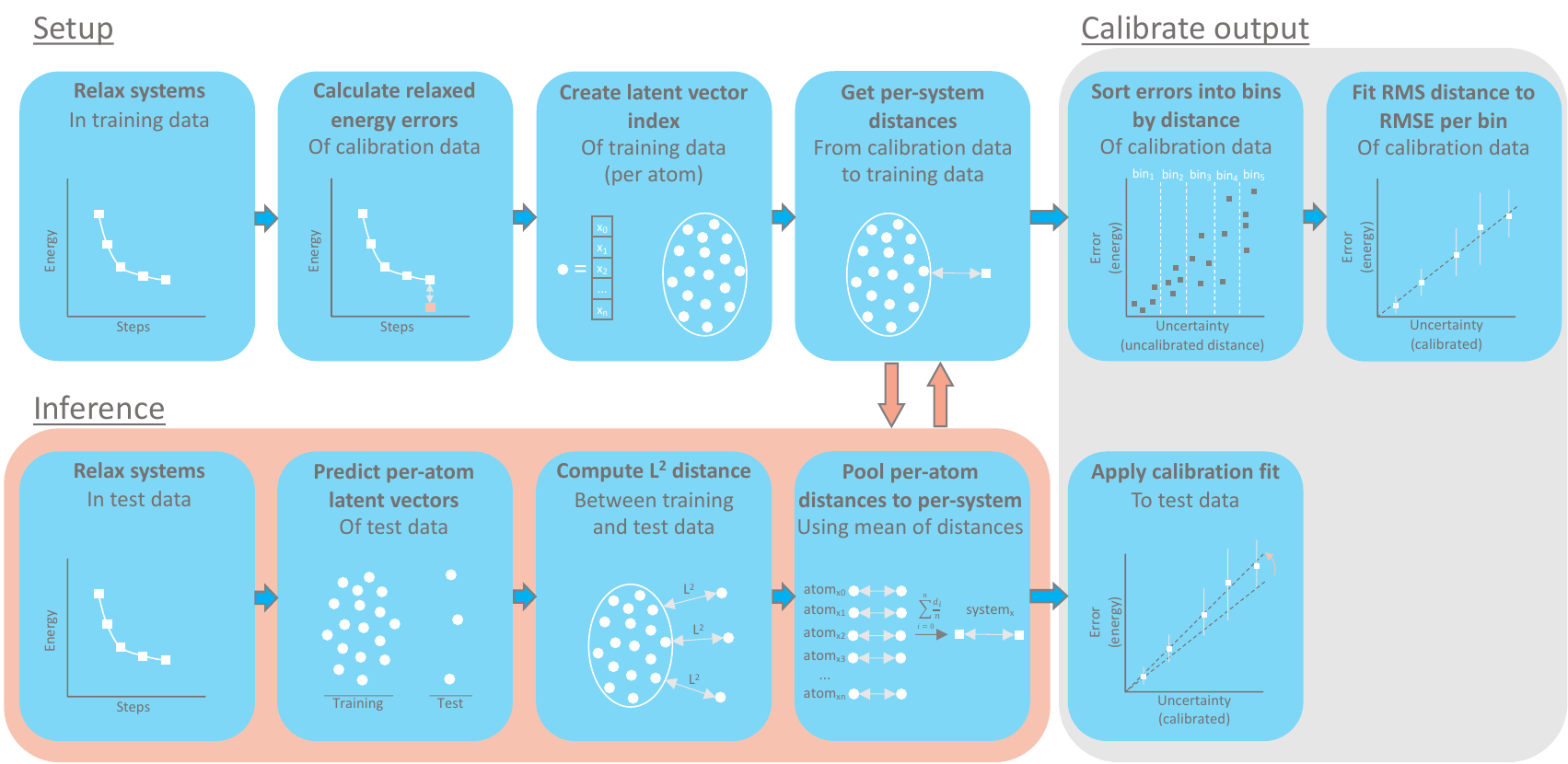}
\caption{Diagram of the components of the latent distance uncertainty prediction method. It details the setup procedure and the inference procedure. The setup procedure must be performed once per model training set. The inference procedure is first used on the calibration set during the setup procedure, and then every time inference is performed afterwards to predict the model uncertainty. Note that the calibration steps at the end of the setup procedure and inference procedure can be applied to any of the uncertainty prediction methods.
}
\label{fig:latent_distance_diagram}
\end{figure}

In this work we examine four common methods of uncertainty prediction on a pre-trained \gls{gnn}: ensembles, latent space distances, mean variance estimation, and sequence regression models. 
The most widely used of these are ensemble methods, where a set of similar surrogate models are trained on similar sets of data to perform the same task, and the variance between their predictions is used to calculate the uncertainty of the model.
We train ensembles of \gls{gnn}s on the \gls{s2ef} task to calculate energies of adsorbate-catalyst structures, but we are interested in the performance of predicting the uncertainty of EquiformerV2 on the relaxed structures.
We construct three different ensembles for testing this approach, which we refer to as architecture (11 members), bootstrap (10 members), and parameter ensembles (6 members).
Prior work has shown that diversity between members is typically the most important factor in developing a well-calibrated and expressive ensemble for uncertainty prediction \cite{survey_uncertainty,ensemble_diversity, trust_ensembles_under_dataset_shift, ensemble_autodiversity}.
The architecture ensemble contains a variety of \gls{gnn} model architectures, while the parameter and bootstrap ensembles contain only EquiformerV2 models, but vary the number of parameters, and the composition of training data respectively. More details on ensemble construction can be found in the supporting information.
Because we are specifically interested in the uncertainty of relaxed energy predictions, which requires a sequence of prior energy/force predictions during the relaxation, we hypothesize that taking the mean or the maximum of the predicted variances over each step (here referred to as a frame) of the trajectory might contain additional information which better models the uncertainty.
For each ensemble composition, we test this theory by computing the uncertainty using the variance at the first frame, last frame, mean over all the frames, and max over over all the frames.

Another proven uncertainty method is the use of latent space distances, commonly referred to as conformal prediction methods \cite{conformal_prediction, conformal_kulik, rasmussen_duan_kulik_jensen_2023}. 
In this approach, we extract some latent representation of each training point from the \gls{gnn}, and create an index of these points to compute the L2-norm of the distance from any new test point to the training points.
In practice, using libraries such as FAISS, the computational cost of this method at inference time is the lowest of any of the uncertainty methods we test \cite{faiss}.
This method also comes with an upfront cost of creating the index of training point representations, but this is generally significantly less expensive than the upfront costs associated with training an additional model or models.
Because this approach produces a distance of arbitrary scale, it is necessary to recalibrate on some calibration set to produce a meaningful uncertainty estimate.
Figure \ref{fig:latent_distance_diagram} shows the pipeline we use to setup, predict uncertainties, and calibrate the uncertainty predictions for this latent distance method. 
Prior work has shown the latent distance method to be effective for models which preserve some rotationally invariant information through the model architecture, such as GemNet-OC. However EquiformerV2 improves upon the accuracy of GemNet-OC by preserving rotational equivariance all the way through the model \cite{equiformerv2,gemnetoc}.
We expect this rotational equivariance will contribute undesirable noise to the L2 distance between latent representations, so we test this hypothesis by comparing the performance of the full EquiformerV2 latent representation, and the latent representation of its single rotationally invariant channel. 
We also compare these methods to the performance of using the latent space representation of other \gls{gnn} models, such as GemNet-OC, trained on the same data, but used to predict the uncertainty of the same EquiformerV2 model.
Additionally, for each latent distance approach, we test a novel strategy of computing the latent distance on a per-atom basis, and then taking the mean, max, or sum over these distances.
We compare this method to the more common approach of computing the latent distance for an entire frame by taking the mean of latent representations over all the atoms.
More details on how all of the latent representations were extracted, and how the distances were computed can be found in the supporting information.

The final categories of methods we test are \gls{mve} and sequence regression models \cite{uq_benchmark_ensembles_good, uq_mve_deep_ensembles, uq_coley_benchmark, meta_sequence_model}. 
We implement these methods in similar ways. 
For \gls{mve} methods we append an output-head, or an ensemble of output-heads to the EquiformerV2 architecture, and fine-tune this new output head on the calibration set to predict the residual of the energy prediction of the larger model. 
For the sequence regression model, we extract the same latent space representations used in the latent distance method, and train a sequence regression transformer architecture to predict the residual of the energy prediction on the calibration set \cite{distilbert}. 
A significant distinction between these approaches is that the sequence regressor takes the latent representations of each frame of the trajectory as input, in sequence, with the hypothesis that some additional information about the uncertainty of the \gls{gnn} model might be contained within its latent representations along the trajectory.
More details on the implementation of both of these approaches can be found in the supporting information.

\subsection{Uncertainty Validation Metrics}
We suspect that the \gls{nll}, Spearman's rank correlation coefficient, and miscalibration area, will not serve as ideal metrics for \gls{uq} validation on this task, due to the non Gaussian distribution of the \gls{rs2re} task. 
We test this by evaluating each of the uncertainty estimation methods using each of the commonly used \gls{uq} validation metrics. 
We anticipate significant disagreement among many of the \gls{uq} validation metrics, especially those not based on distribution-free methods. 
Further examination of the error distribution of the \gls{rs2re} task can be found in the supporting information.

Prior work by Pernot suggests the use of a distribution-free method to test whether an uncertainty method is calibrated \cite{pernot}.
The CI(Var(Z)) test uses the BCa boostrap method to compute a confidence interval of Var(Z) on a set of errors and uncertainties, without making any assumptions about the distribution \cite{bca_bootstrap, scipy_bootstrap}.
If 1 lies within the confidence interval, the uncertainty method is considered calibrated, because the Z values for a calibrated uncertainty metric are expected to have a variance of 1.
When applied to the calibration set, this metric is useful for characterizing whether error-based recalibration is effective in achieving globally well calibrated uncertainties.
On the test set (after recalibration on the calibration set) estimated uncertainties are typically no longer perfectly calibrated. 
However the closer the confidence interval of the variance is to the ideal (Var(Z)=1), the better the calibrated uncertainty estimates remain globally calibrated under distribution shift.
Unless otherwise stated, we report the CI(Var(Z)) metric on the test set.

Rasmussen et al. expands on this to suggest using an error-based calibration plot to quantify local calibration, distribution-free \cite{levi2022_error_based_cal, rasmussen_duan_kulik_jensen_2023}.
The error-based calibration plot is predicated on the expected relationship between the \gls{rmv} and \gls{rmse} being one-to-one.
\begin{equation}
    RMSE = \sqrt{1/N_{bin}\sum_{i}\epsilon^{2}_{i}} \quad RMV = \sqrt{1/N_{bin}\sum_{i}\sigma^{2}_{i}} \quad \frac{RMSE}{RMV} \approx 1
\end{equation}
We sort the test points by their predicted uncertainty, and then bin them into 20 bins.
We compute the \gls{rmv} of each bin, and the \gls{rmse} of each bin, using the BCa bootstrap method to compute a 95\% confidence interval for the \gls{rmse}.
Then we fit a line through the points to test for calibration, ideally the fitted line should have a high \rsq\ correlation with the points, and be as close as possible to the parity line.
We can identify problems with local miscalibration where the parity line does not lie within the binned \gls{rmse} confidence intervals.

\subsection{Recalibration and Evaluation}
\label{sec:recalibration}
We compare all of the uncertainty methods benchmarked in this work after recalibration on a calibration set.
In this case we use \gls{oc20} in-domain validation set, and relax each structure (approximately 25,000 structures) to the same relaxation criterion as \gls{oc20}, with the publicly available EquiformerV2 31M parameter checkpoint.
We then compute a single point calculation using \gls{vasp} on the final frame to serve as the ground truth energy value for the RS2RE task \cite{vasp1,vasp2,vasp3,vasp4}.
This set of relaxed energy predictions serves as the calibration set, and we repeat this process with the out-of-domain-both validation set to serve as the test set for this task.
To recalibrate each uncertainty method, we use same the approach as the error-based calibration plot, to find the line of best fit through the binned \gls{rmse}/\gls{rmv} points of the calibration set.
We then simply recalibrate the uncertainty using the formula for the line of best fit:
\begin{equation}
    \sigma_{recalibrated} = \text{slope}_{fit}*\sigma_{uncalibrated} + \text{intercept}_{fit}
\end{equation}
We propose a small modification to the characterization of global calibration of uncertainties recalibrated with the error-based method.
For this task, in addition to checking the global calibration CI(Var(Z)) test, we also select the most effective uncertainty metric by measuring the $R^{2}$ correlation of the binned \gls{rmse}/\gls{rmv} ratio with the parity line.
By recalibrating the uncertainty method on the calibration set and then testing on the test set, we can make direct comparisons between different uncertainty methods on the basis of their $R^{2}$ correlation with the parity line.

\section{Results}

\subsection{Error Distribution and Uncertainty Validation Metrics}

\begin{figure}
% -------------subfig 1-------------
 \begin{subfigure}{0.47\textwidth}
     \includegraphics[width=\textwidth]{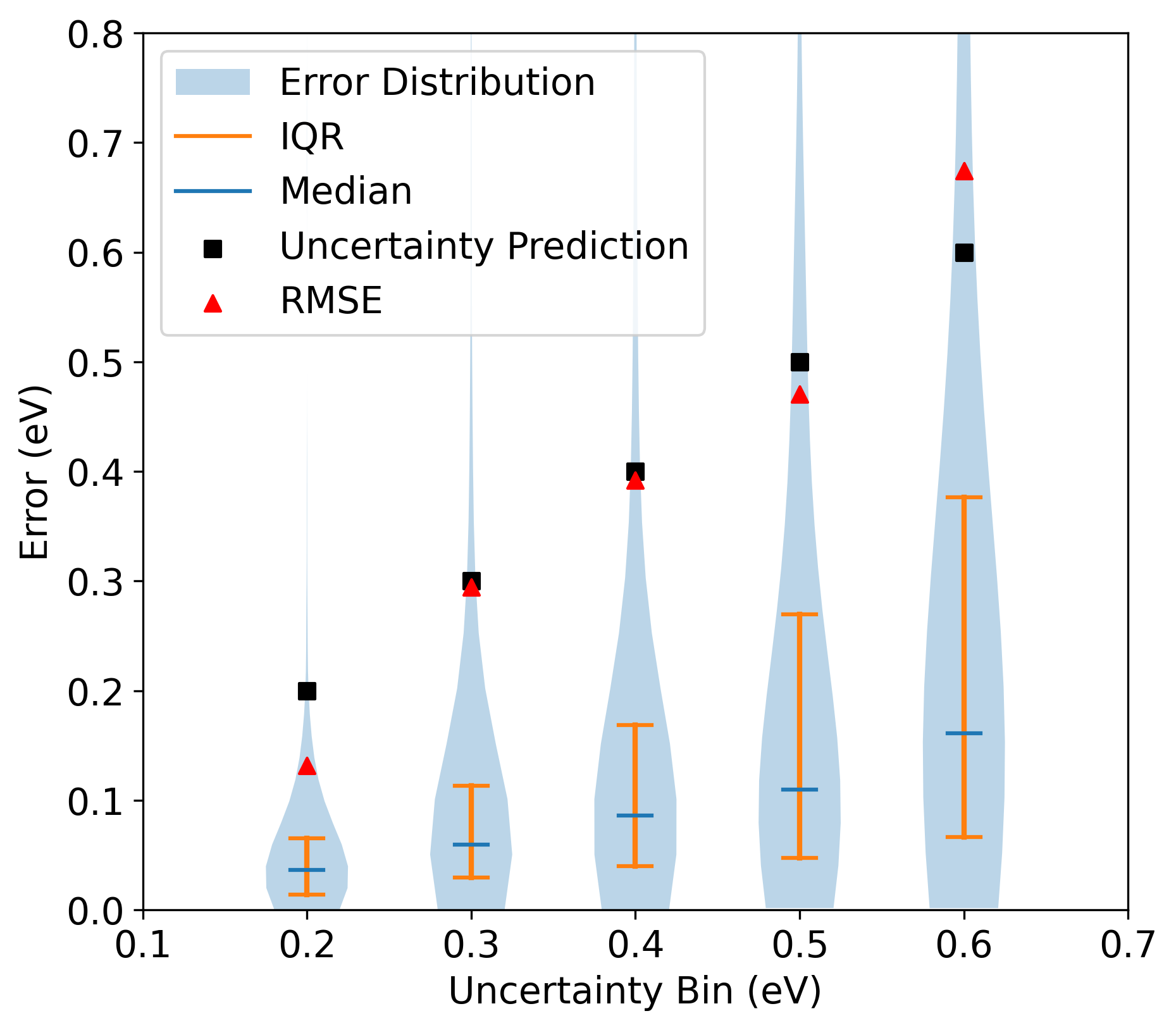}
     \caption{}
     \label{fig:violin}
 \end{subfigure}
% -------------subfig 2-------------
 \begin{subfigure}{0.5\textwidth}
     \includegraphics[width=\textwidth]{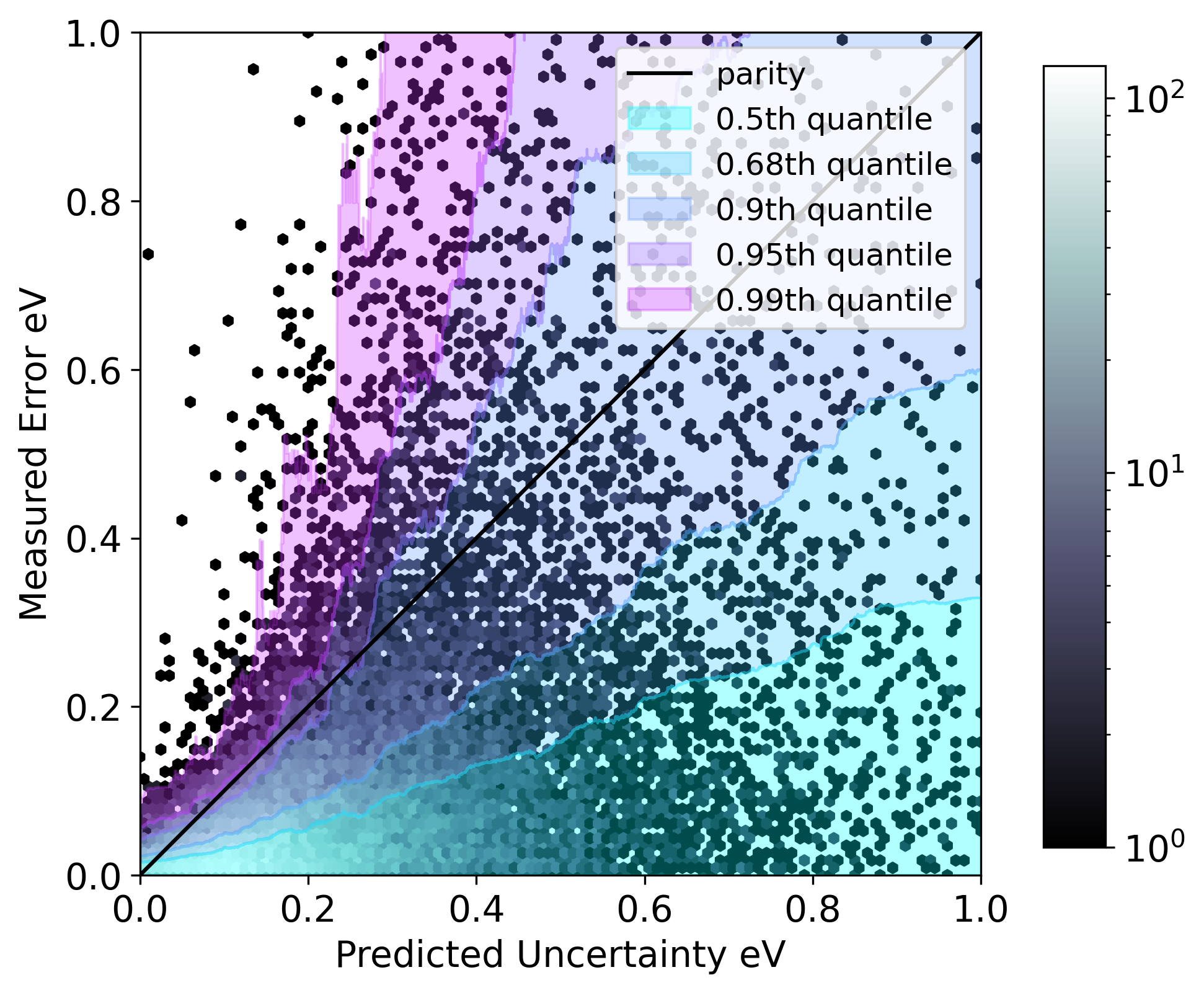}
     \caption{}
     \label{fig:contour}
 \end{subfigure}

\caption{
    Violin and parity plot for a latent distance uncertainty prediction. After calibrating uncertainty predictions with error based calibration, they can be interpreted as predictions of the dispersion of the residual for that inference calculation. In \ref{fig:violin} we see the measured error distributions for five uncertainty prediction bins. In \ref{fig:contour} we see contour lines tracing five different quantiles of measured error, rolling along the predicted uncertainty.
}
\label{fig:contour_violin}
\end{figure}

To interpret the meaning of these recalibrated uncertainties, we look at the distribution of errors compared to the corresponding uncertainties. 
In Figure \ref{fig:violin} we plot the error distributions of points in the test set binned by their estimated uncertainties, after error-based recalibration. 
When we recalibrate uncertainty predictions, we fit the ratio of the \gls{rmse} to the \gls{rmv} of these bins to the parity line. 
We see that the distribution of errors is generally skewed, with the bulk of the distribution falling below the predicted \gls{rmv} (uncertainty) of the bin. 
This is what we expect to see for uncertainty predictions, since calibrated uncertainties should describe the dispersion of the distribution that the error is sampled from, most of the sampled errors should lie below the predicted uncertainty.
We also see that the measured \gls{rmse} generally corresponds to the predicted \gls{rmv} of each bin. 
However the lowest uncertainty bin is underconfident, and the highest uncertainty bin is overconfident, which is a common trend across many uncertainty estimators we examine in this work.

In the parity plot in Figure \ref{fig:contour} we see the effect of the error-based recalibration on the distribution of \gls{rs2re} errors. 
The contour lines trace the n-th quantile of the measured errors. 
The contour lines curve somewhat relative to the parity line, this is partly explained by the underconfidence towards the lower end of the plot, and overconfidence towards the highest uncertainties. 
However this is also an effect of the error distribution being non-Gaussian.
We do not necessarily expect to see a perfect linear relationship between the contour lines and the parity line, unless the errors follow a Gaussian distribution, since calibration is only based on a linear relationship between \gls{rmse}/\gls{rmv}. 
Even with a non-Gaussian distribution, the uncertainties can still be well-calibrated with respect to the errors, as long as the \gls{rmse}/\gls{rmv} relationship follows the parity line. 
The goal of a well-calibrated uncertainty estimate is to predict the dispersion characteristic of the error distribution at that point, regardless of the distribution from which it is sampled.
If, for instance, the errors more closely follow a Laplace distribution, we would expect to see contour lines similar to the ones we observe here.

\begin{table}
\caption{Best performers for each of the \gls{uq} validation metrics. Many of the popular \gls{uq} validation metrics assume normally distributed errors, such as miscalibration area ($A_{mis}$), before and after error-based recalibration, and the negative log likelihood (NLL). Other metrics target properties other than calibration, such as \spearman ($\rho$) and the area under the receiver operating characteristic curve (AUROC). These methods tend to be inconsistent with one another. The error based calibration (Parity $R^{2}$) measure and the CI(Var(Z)) test both measure calibration without a distribution assumption, and agree that the GNOC distance method performs best.}
\label{tab:uq_top_per_metric}
\centering

\begin{tabular}{lrrrrrrl}
\toprule
 Method            &   Parity $R^{2}$ &   $A_{mis,u}$ &   $A_{mis,r}$ &   $NLL$ &   $\rho$ &   AUROC & CI(Var(Z))   \\
\midrule
 GNOC distance     &            \textbf{0.952} &         0.365 &         0.178 &   0.279 &    0.468 &   0.755 & \textbf{[1.13, 1.79]} \\
 Architecture ens. &            0.830 &         \textbf{0.050} &         0.156 &   0.411 &    0.453 &   0.746 & [1.58, 2.19] \\
 Residual model    &            0.331 &         0.174 &         \textbf{0.102} &   1.170 &    0.360 &   0.695 & [3.38, 4.47] \\
 MVE head          &            0.831 &         0.340 &         0.114 &   \textbf{0.251} &    0.526 &   0.785 & [1.59, 2.40] \\
 Bootstrap ens.    &            0.870 &         0.248 &         0.121 &   0.303 &    \textbf{0.585} &   \textbf{0.817} & [1.58, 2.48] \\
\bottomrule
\end{tabular}

\end{table}

In Table \ref{tab:uq_top_per_metric} we compare the best performing \gls{uq} methods according to several commonly used \gls{uq} validation metrics. 
We observe that top performance in one metric does not correspond to top performance in another, except with respect to the distribution-free validation metrics, Parity $R^{2}$ and the CI(Var(Z)), which tend to agree.
Unsurprisingly, \spearman and AUROC also agree, as both of these metrics are aimed at classification performance, but do not provide much insight into overall calibration. 
These results indicate that the distribution-free validation metrics output significantly different results from other common \gls{uq} validation metrics, and that other such metrics cannot be relied upon to agree with one another.
Due to the fragility of assumptions of Gaussian distributed errors for the \gls{rs2re} task, and the recommendations from previous work on this topic, we are inclined to believe that these distribution-free metrics are more reliable than other options for generally assessing and comparing calibration on the \gls{rs2re} task \cite{pernot, levi2022_error_based_cal, rasmussen_duan_kulik_jensen_2023}.

\subsection{Benchmarking Uncertainty Quantification}

\begin{table}
\caption{Distribution free calibration metrics for the best performing candidate uncertainty estimation method for each of the categories we benchmark. Note that all of the uncertainty estimates are able to be globally calibrated on the calibration set. The latent space distance method achieves highest local calibration, and is closest to well calibrated globally on the test set, according to the CI(Var(Z)) test.}
\label{tab:benchmark}
\centering

\begin{tabular}{lrrrrll}
\toprule
 Method          &   Parity $R^{2}$ &   Fit $R^{2}$ &   Slope &   Int. & Cal. set      & Test set     \\
                 &                  &               &         &        & CI(Var(Z))    & CI(Var(Z))   \\
\midrule
 Latent distance &   \textbf{0.952} &         0.967 &   1.022 &  0.026 & [0.84, 1.35]* & [1.13, 1.79] \\
 Ensemble        &            0.905 &         0.940 &   0.946 &  0.070 & [0.83, 1.57]* & [1.42, 2.57] \\
 Residual model  &            0.331 &         0.955 &   1.200 &  0.127 & [0.89, 1.50]* & [3.38, 4.47] \\
 MVE head        &            0.849 &         0.964 &   1.187 &  0.041 & [0.86, 2.09]* & [1.66, 2.46] \\
\bottomrule
\end{tabular}

\end{table}

Of the four methods we benchmark on the \gls{rs2re} task, the latent distance method is the best performer according to the distribution-free uncertainty quantification techniques, as seen in Table \ref{tab:benchmark}.
The CI(Var(Z)) global calibration test shows that all four \gls{uq} methods seem to be well calibrated by error based recalibration on the calibration set.
However on the test set, all of the \gls{uq} some degree of miscalibration due to distribution shift.
The latent distance approach demonstrates the best test set performance, according to the CI(Var(Z)), since the confidence interval is closest to 1. 
It also achieves the best calibration according to the parity \rsq\ metric on the out-of-domain test set.
Note that all four models are capable of a comparable fit (if this test set were used for recalibration) as evidenced by the similar scores on the fit \rsq.

\begin{figure}
\includegraphics[width=\textwidth]{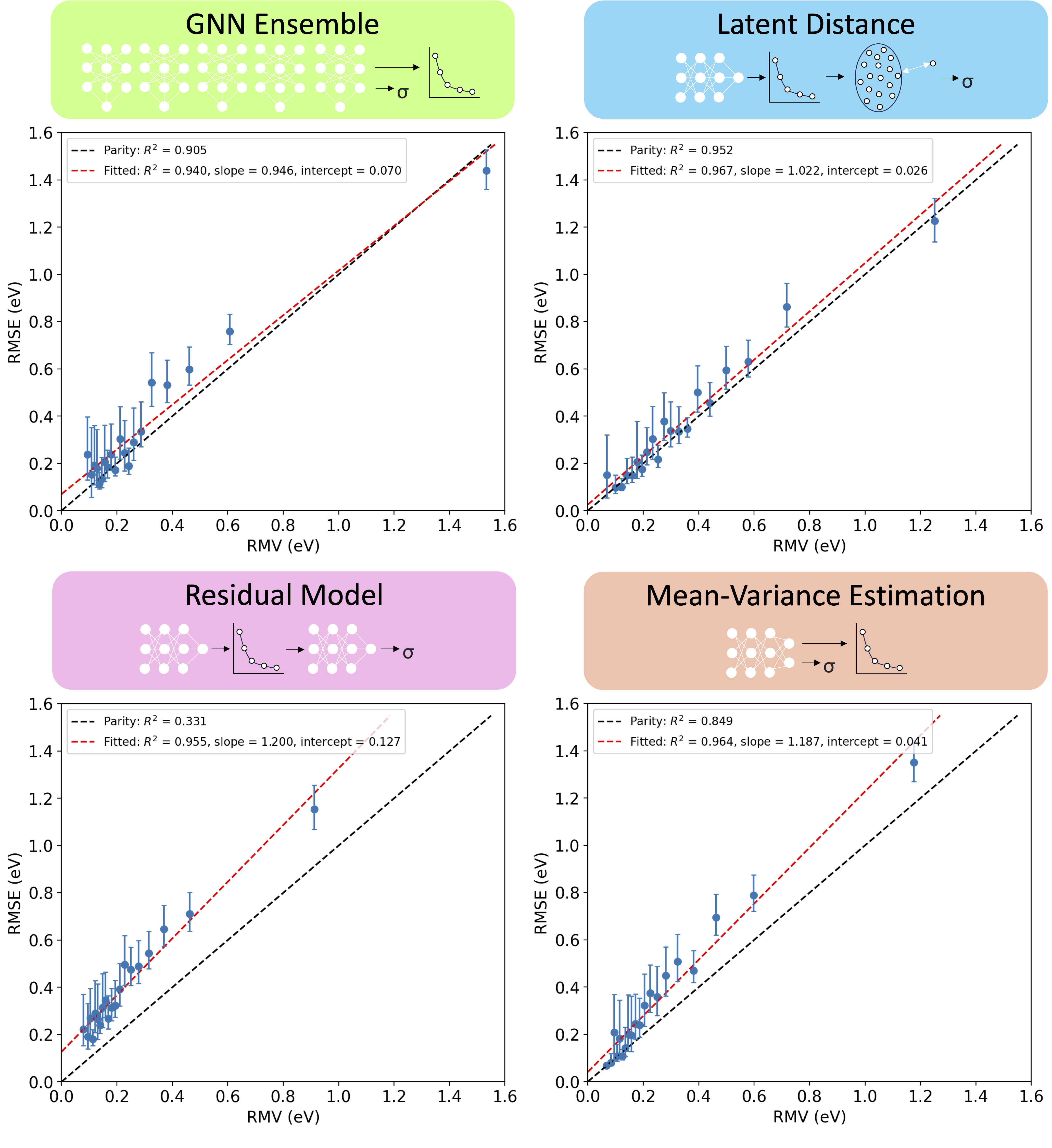}
 \caption{Calibration plots for the best performing iterations of the four categories of methods discussed in this work. The diagram above each plot shows how each method estimates the variance of the relaxed energy trajectory predicted by the graph neural network.}
 \label{fig:methods_diagram}
\end{figure}

We observe superior local calibration for the top performing latent distance method, as compared to the other \gls{uq} methods in Figure \ref{fig:methods_diagram}.
We see that the latent distance method generally retains good local calibration throughout, which we can see by the error bars of nearly every confidence interval intersect the parity line.
The latent distance method also maintains the best overall calibration, with the line of best fit on the the test set only being offset slightly from the parity line.
The ensemble method, which is the next best method class, sees more significant local deviations, displaying intermittent overconfidence below 0.3 eV, then systematic overconfidence between 0.3 eV and 1 eV. 
The ensemble method does maintain a good overall calibration fit on the test set.
The remaining two methods both result in more significant deviations on the test set.
The residual model is systematically over-confident on the entire test set, and this shows in its overall parity \rsq.
The \gls{mve} method only achieves good local calibration at the lowest uncertainty estimates, before becoming systematically overconfident above 0.2 eV.

\begin{figure}
 \begin{subfigure}{0.49\textwidth}
     \includegraphics[width=\textwidth]{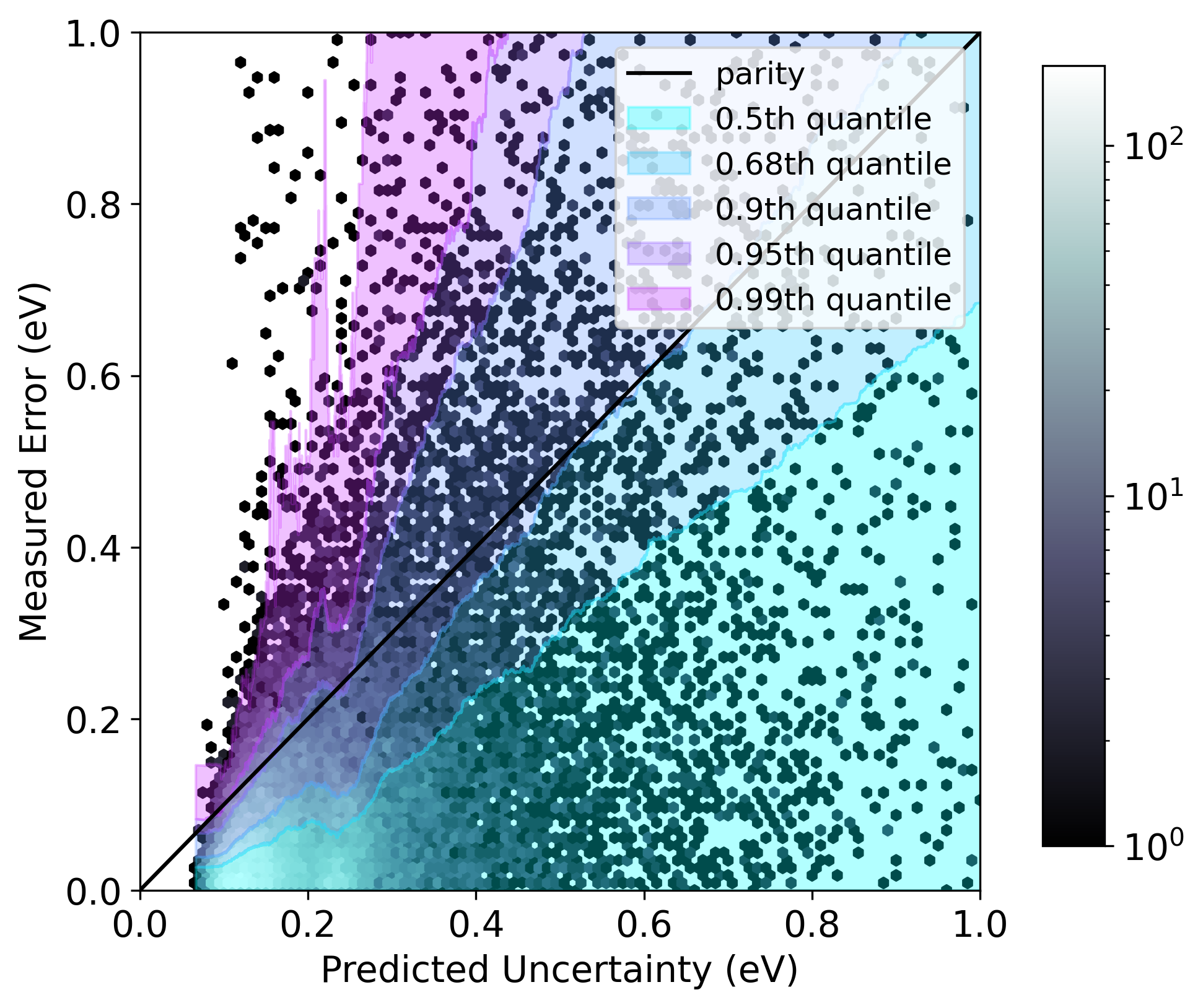}
     \caption{Best ensemble contour plot \\
     (Architecture ensemble, mean over trajectory)
     }
     \label{fig:best_ens_contour}
 \end{subfigure}
 \begin{subfigure}{0.49\textwidth}
     \includegraphics[width=\textwidth]{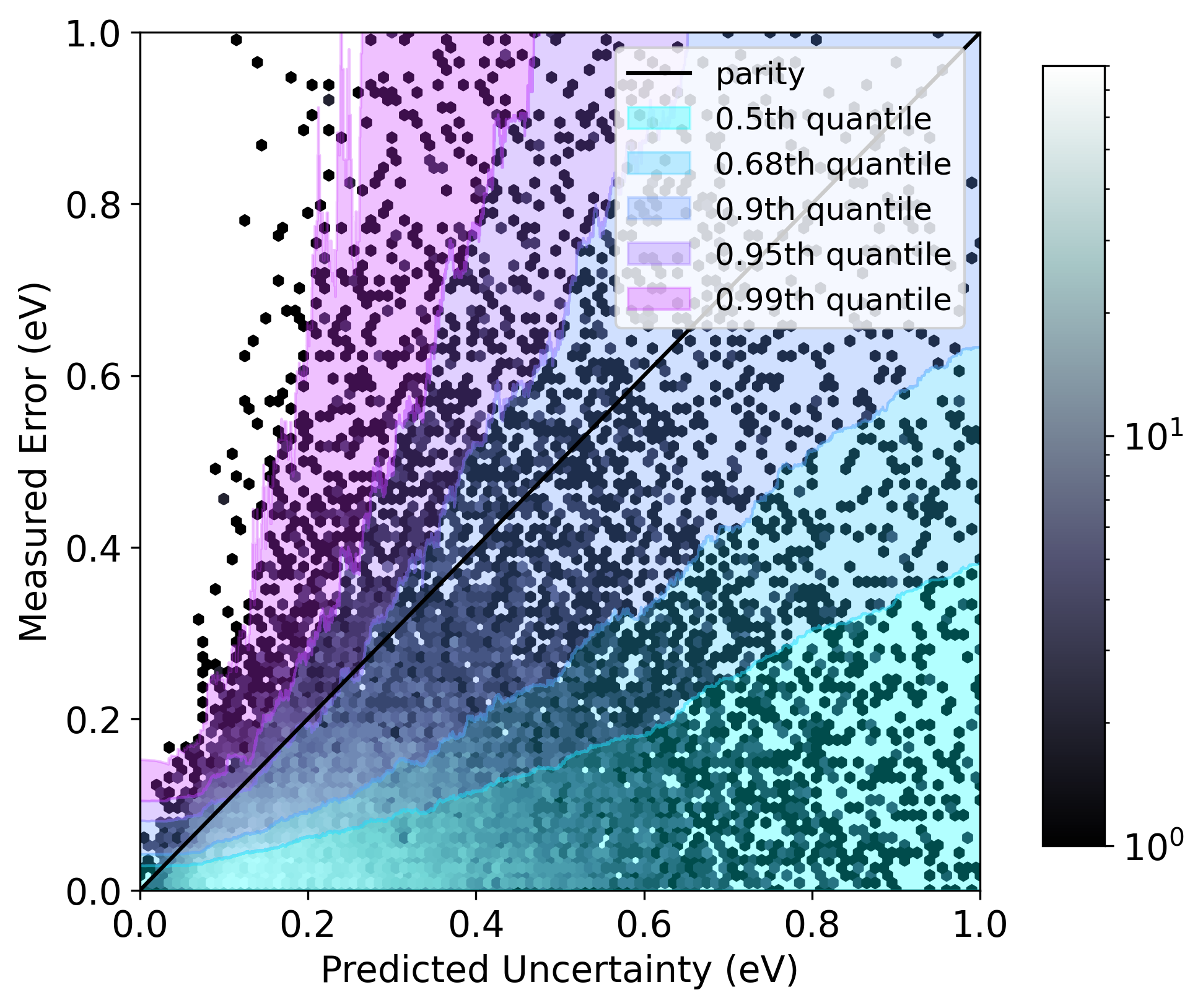}
     \caption{Best latent distance contour plot \\
     (GemNet-OC, sum of per-atom distances)
     }
     \label{fig:best_dis_contour}
 \end{subfigure}

 \caption{Contour parity plots for the best two method classes, predicting on the test set after recalibration on the calibration set. Uncertainties predicted by the best performing ensemble method (\ref{fig:best_ens_contour}) and the best performing latent distance method (\ref{fig:best_dis_contour}). The quantile curves of the latent distance method are generally montonic, which aligns with the better local calibration performance. The ensemble method uncertainties are shifted right by recalibration, and shows a significant drop in the uncertainty estimates after 0.2 eV, and an abrupt increase after 0.3 eV.}
 \label{fig:ensemble_vs_distance}
\end{figure}

We examine the contour parity plots of the two top performing \gls{uq} methods in Figure \ref{fig:ensemble_vs_distance}.
The ensemble method shows a shift which lower bounds the predicted uncertainties resulting from recalibration, while the recalibrated uncertainties of the latent distance method result in no such lower bound.
This lower bounding limits the effectiveness of this approach at low uncertainty estimates, which may be a critical region of interest depending on the use case.
We also observe a sharp drop and increase in the measured error of the ensemble method from 0.2 eV to 0.3 eV, which seems to be responsible for the intermittent local miscalibration in that range.
Overall, we see the error distribution predicted by the latent distance method fans out in a more linear fashion.
The parity line appears to be a rough upper bound for the errors for the latent distance method, which is desirable in a well-calibrated uncertainty method.

\subsection{Comparing Distance Methods}

The latent space representation sampled from EquiformerV2 after all graph convolutional interactions have been performed is inherently equivariant with respect to rotations.
This is a valuable property for training \gls{gnn} machine learning potentials to predict properties of molecular systems, but it may contribute noise to distance measures in the latent space.
Since the latent space distance should measure the similarity of atomic structure representations, it should ideally be invariant to the rotations of input structures.
Since the same structure can be rotated to a numerically different latent space representation, many more samples must be present in training set for the latent space distance to be meaningful.
Therefore we compare distances measured using the full latent space representation, which includes all spherical harmonic channels, to distances measured using only the rotationally invariant (degree 0 spherical harmonic channels) latent space representations.

\begin{figure}
 \begin{subfigure}{0.49\textwidth}
     \includegraphics[width=\textwidth]{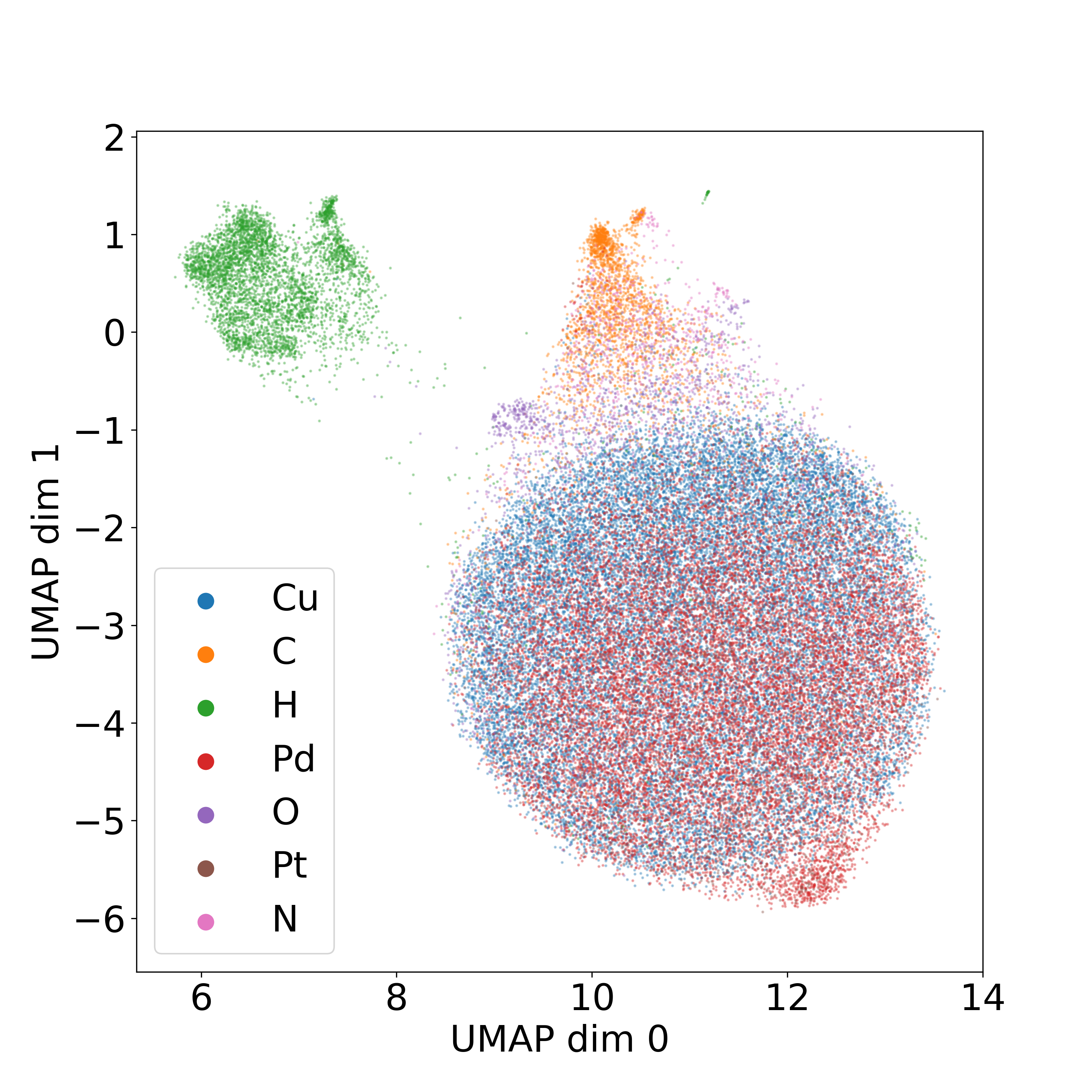}
     \caption{Equivariant Latent Space}
     \label{fig:equiv_umap}
 \end{subfigure}
 \hfill
 \begin{subfigure}{0.49\textwidth}
     \includegraphics[width=\textwidth]{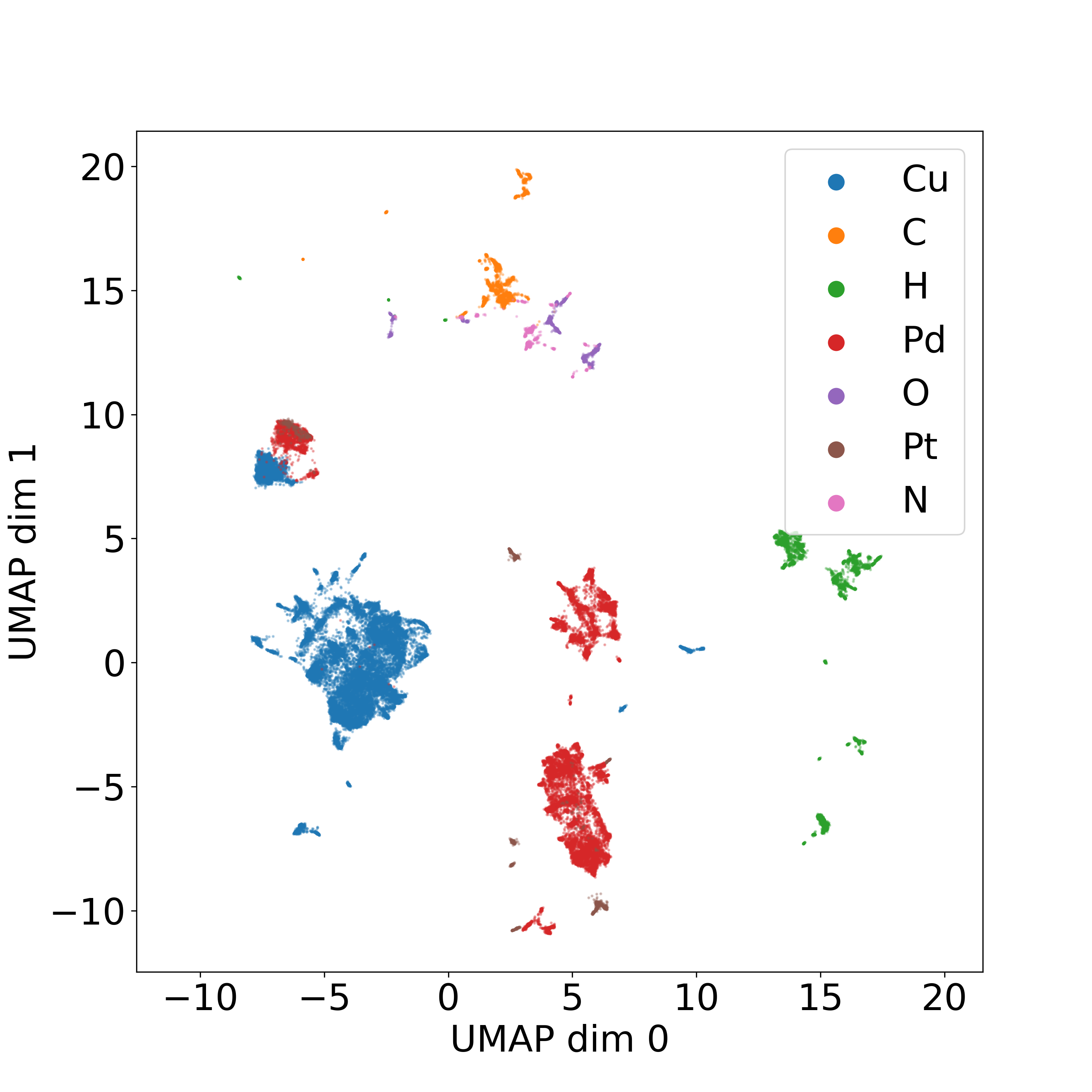}
     \caption{Invariant Latent Space}
     \label{fig:inv_umap}
 \end{subfigure}
 \caption{Plots of UMAP dimensionality reduction performed on the equivariant (all channels) and invariant (m=0, l=0) latent space representations sampled from Equiformer V2 without edge alignment. Latent space representations were sampled for a subset of the training set containing specific elements. We see that the different elements represented are clearly clustered in both plots, but that there is significantly more noise in the clustering of similar elements in the equivariant latent space, while the invariant latent space clusters are much denser and less noisy.}
 \label{fig:umap}

\end{figure}

In Figure \ref{fig:umap} we observe the effects of these two different latent space sampling methods on Equiformer V2 using UMAP, a dimensionality reduction package \cite{umap}. 
We plot the first two dimensions of the UMAP-reduced latent space, sampled from a subset of systems containing only three metal species, and four adsorbate species. 
We observe clear clustering of the latent space representations based on the atomic species, but with significantly more noise present in the fully equivariant latent representation. 
By contrast, the invariant latent representation produces much denser clusters, which are more clearly differentiated by atomic species.
The key features we would expect to see in a good atomic representation are: clear distinctions between metals and adsorbate atoms, and more similarity between group 10 metals like Pd and Pt than between those and group 11 metals like Cu.
Both latent representations produce definitive segregation of adsorbate species from metal species, however the equivariant representation generates a cloud of overlapping metal species which includes Cu, Pd, Pt.
The invariant representation creates distinct clusters for Cu, and clusters with significant overlap between Pt and Pd.
This seems to indicate that the rotationally equivariant information, which is preserved by the structure of Equiformer V2, will appear as noise to simple low dimensional comparisons of latent representations.
This explanation also aligns with the superior performance of distance measures within rotationally invariant latent space.

\begin{table}
\caption{
Distribution free calibration metrics for each latent distance method tested. The latent representations correspond to model and method that was used to extract the latent representation for each atom. The distance column refers to the approach used to compute the distance between systems, which are described in the methods section. A * indicates the method is globally calibrated according the CI(Var(Z)) test.
}
\label{tab:distances}
\centering

\begin{tabular}{llrrrll}
\toprule
 Latent Rep.           & Distance   &   Parity $R^{2}$ &   Slope &   Int. & Cal. set      & Test set      \\
                       &            &                  &         &        & CI(Var(Z))    & CI(Var(Z))    \\
\midrule
 EqV2 equiv.           & atom max   &           -0.078 &   0.761 &  0.182 & [0.90, 1.16]* & [1.61, 2.02]  \\
 EqV2 equiv.           & atom mean  &            0.813 &   0.974 &  0.060 & [1.03, 1.74]  & [1.51, 5.45]  \\
 EqV2 equiv.           & sys. mean  &          -14.709 &  -0.236 &  0.547 & [1.24, 2.74]  & [1.90, 3.93]  \\
 EqV2 equiv.           & atom sum   &            0.672 &   0.711 &  0.159 & [0.89, 1.27]* & [1.52, 2.78]  \\
 EqV2 inv. l=0 (nodes) & atom max   &            0.842 &   0.983 &  0.054 & [0.86, 1.24]* & [1.22, 1.70]  \\
 EqV2 inv. l=0 (nodes) & atom mean  &            0.866 &   0.974 & -0.037 & [1.16, 3.10]  & [1.04, 5.44]  \\
 EqV2 inv. l=0 (nodes) & sys. mean  &           -0.628 &   1.540 & -0.061 & [0.91, 1.14]* & [1.69, 2.05]  \\
 EqV2 inv. l=0 (nodes) & atom sum   &            0.826 &   0.750 &  0.081 & [0.90, 1.47]* & [0.93, 1.59]* \\
 EqV2 inv. l=0 (edges) & atom max   &            0.548 &   1.124 &  0.057 & [0.94, 1.23]* & [1.61, 2.00]  \\
 EqV2 inv. l=0 (edges) & atom mean  &            0.910 &   1.014 & -0.002 & [0.94, 1.65]* & [1.34, 6.37]  \\
 EqV2 inv. l=0 (edges) & sys. mean  &            0.808 &   1.107 &  0.021 & [0.91, 1.14]* & [1.25, 1.55]  \\
 EqV2 inv. l=0 (edges) & atom sum   &            0.882 &   0.814 &  0.094 & [0.88, 1.25]* & [1.14, 1.59]  \\
 EqV2 inv. l=4 (edges) & atom max   &            0.480 &   1.011 &  0.082 & [0.95, 1.22]* & [1.54, 1.88]  \\
 EqV2 inv. l=4 (edges) & atom mean  &            0.906 &   0.998 & -0.020 & [1.00, 2.33]  & [1.10, 5.16]  \\
 EqV2 inv. l=4 (edges) & sys. mean  &            0.824 &   1.121 &  0.006 & [0.90, 1.15]* & [1.19, 1.48]  \\
 EqV2 inv. l=4 (edges) & atom sum   &            0.828 &   0.740 &  0.097 & [0.90, 1.38]* & [1.00, 1.63]* \\
 GNOC                  & atom max   &            0.924 &   0.951 &  0.064 & [0.81, 1.17]* & [1.23, 1.73]  \\
 GNOC                  & atom mean  &            0.932 &   1.108 &  0.006 & [0.84, 1.55]* & [1.17, 2.18]  \\
 GNOC                  & sys. mean  &            0.818 &   1.126 &  0.050 & [0.87, 1.16]* & [1.56, 2.04]  \\
 GNOC                  & atom sum   &   \textbf{0.952} &   1.022 &  0.026 & [0.84, 1.35]* & [1.13, 1.79]  \\
\bottomrule
\end{tabular}

\end{table}

Table \ref{tab:distances} shows that the choice of latent space representation seems to be the most important factor in choosing a good distance metric for predicting uncertainty.
Depending on the approach to sampling an invariant latent space, the performance of using the latent space distance can vary wildly.
These results suggest the second most important factor is choosing a per-atom distance metric, instead of taking the mean of the representations over all the atoms.
Using a distance approach for latent representations pooled over the entire can work passably well in some cases, such as for GemNet-OC, but generally a per-atom approach performs better.
Finally, selecting the right type of per atom distance measure appears to make some difference, but seems to be less significant if a good latent representation is chosen.
Taking the sum or the mean of the per-atom distances tends to work well for most methods shown here.

We also consider how edge alignment impacts the latent space representation during graph convolutions within Equiformer V2.
Atoms in Equiformer V2 are embedded as nodes, with edge-degree embeddings, during equivariant graph attention blocks, the edge embeddings are rotated to align with the y axis and processed by the Transformer block. 
The edge aligned embeddings should be invariant with respect to rotation along all of the m=0 spherical harmonic channels, so we test this by sampling these edge embeddings and pooling them to obtain rotationally invariant per-atom edge embeddings for all of the channels through [l=4,m=0].
We compare this approach to the previously mentioned approach of sampling the rotationally invariant [l=0, m=0] node embedding as a latent representation.
We also compare to using the [l=0, m=0] edge aligned embedding as an invariant latent representation, and to using the GemNet-OC node embeddings as a latent representation, since GemNet-OC inherently maintains rotational invariance through its latent space.
We find that GemNet-OC performs best of these of these approaches, with both of the edge aligned embedding approaches coming in close second, ahead of simply taking the Equiformer V2 [l=0, m=0] node embeddings.
There is little discernible difference between the two edge aligned embedding approaches for Equiformer V2, perhaps indicating that a significant portion of the relevant information is contained in the [l=0, m=0] channel.
The fact that both edge embedding approaches outperform the node embedding approach also may suggest that there is some advantage to using the edge embeddings for distance metrics, that goes beyond eliminating noise from rotational equivariance. 
All of these approaches far exceed the performance of using all the equivariant spherical harmonic node embeddings.

\subsection{Interpretable Examples}

\begin{figure}
% -------------subfig 1-------------
 \begin{subfigure}{0.33\textwidth}
     \includegraphics[width=\textwidth]{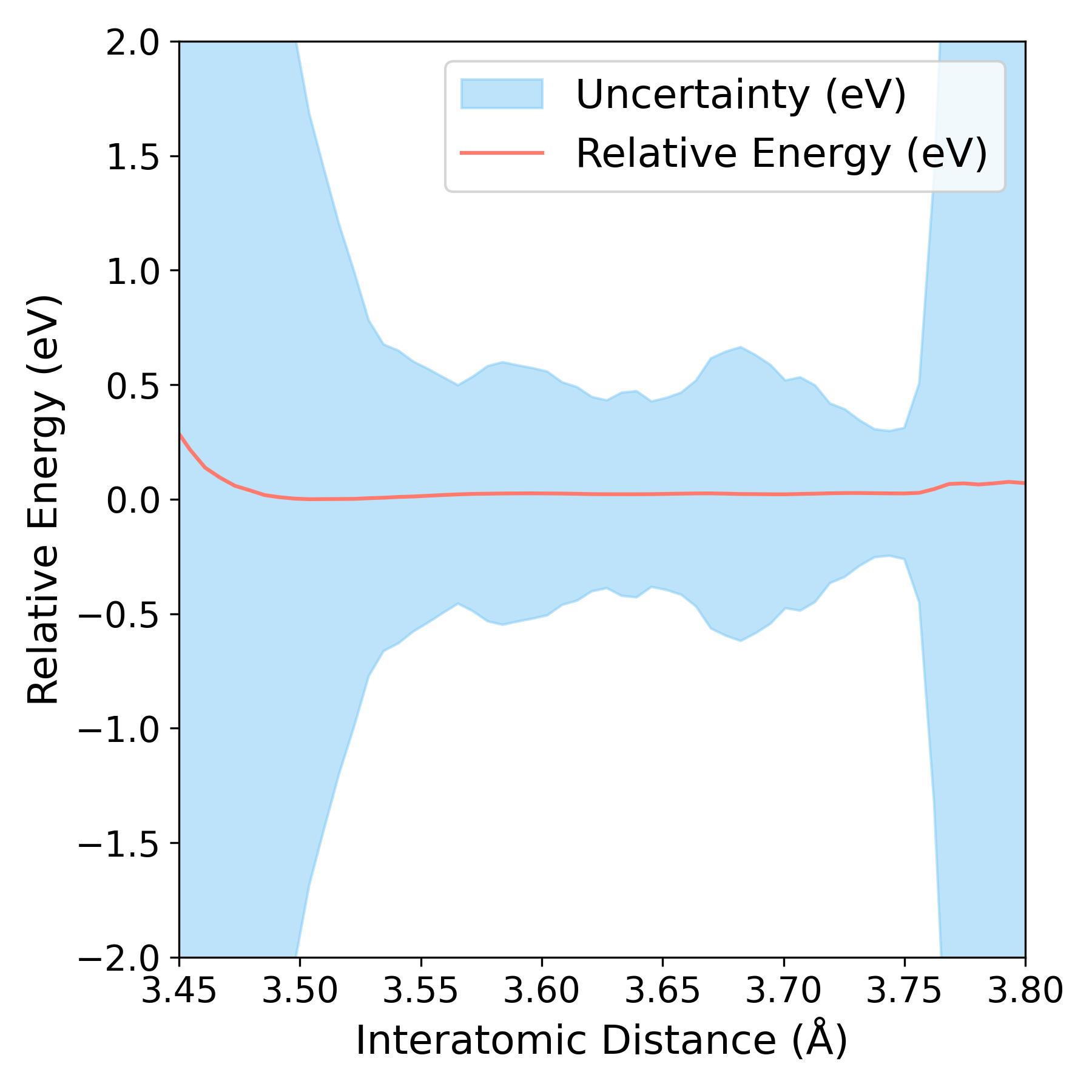}
     \caption{
        Cu bulk
     }
     \label{fig:eos_cu}
 \end{subfigure}
% -------------subfig 2-------------
 \begin{subfigure}{0.33\textwidth}
     \includegraphics[width=\textwidth]{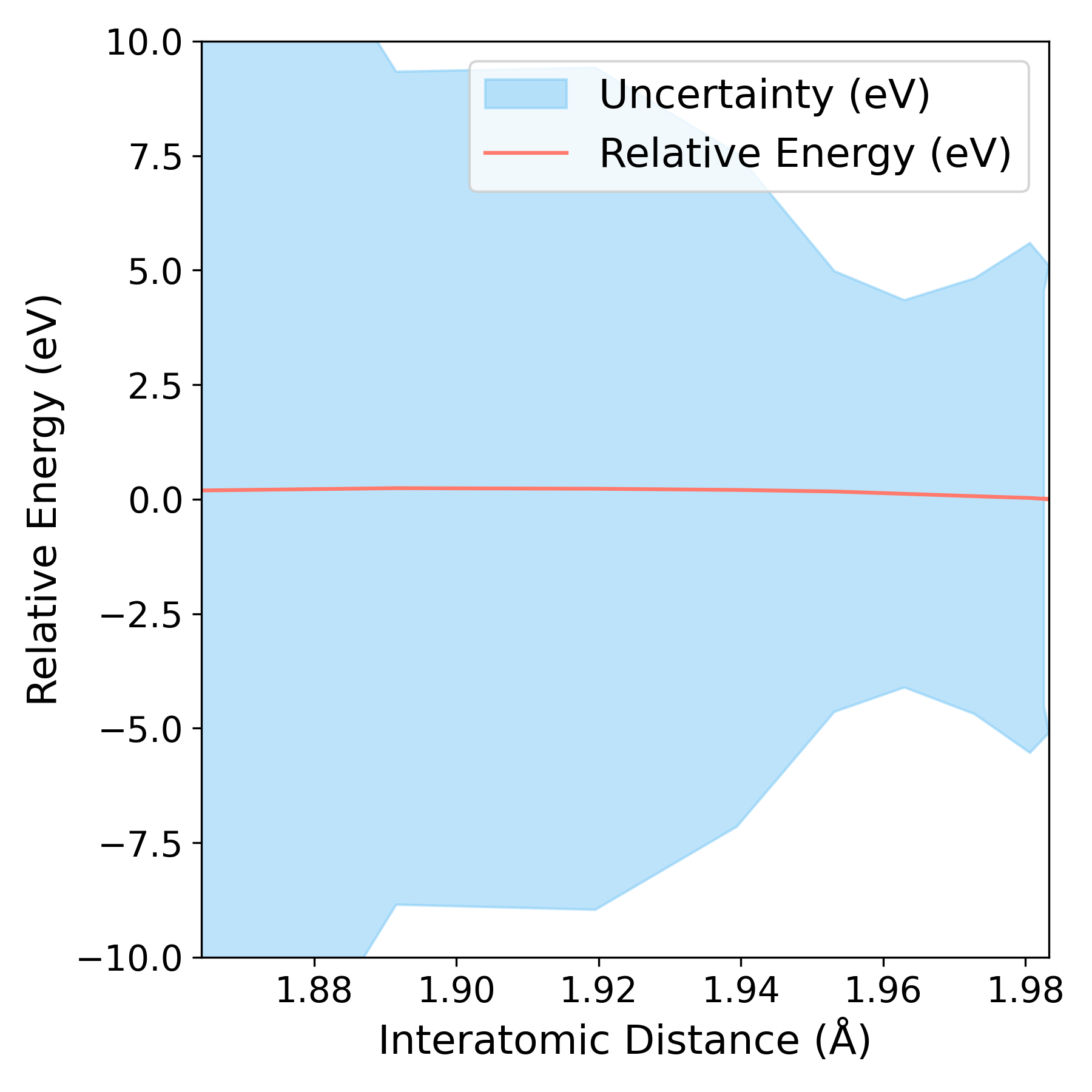}
     \caption{
        TiO\textsubscript{2} bulk
     }
     \label{fig:eos_tio2}
 \end{subfigure}
% -------------subfig 3-------------
 \begin{subfigure}{0.33\textwidth}
     \includegraphics[width=\textwidth]{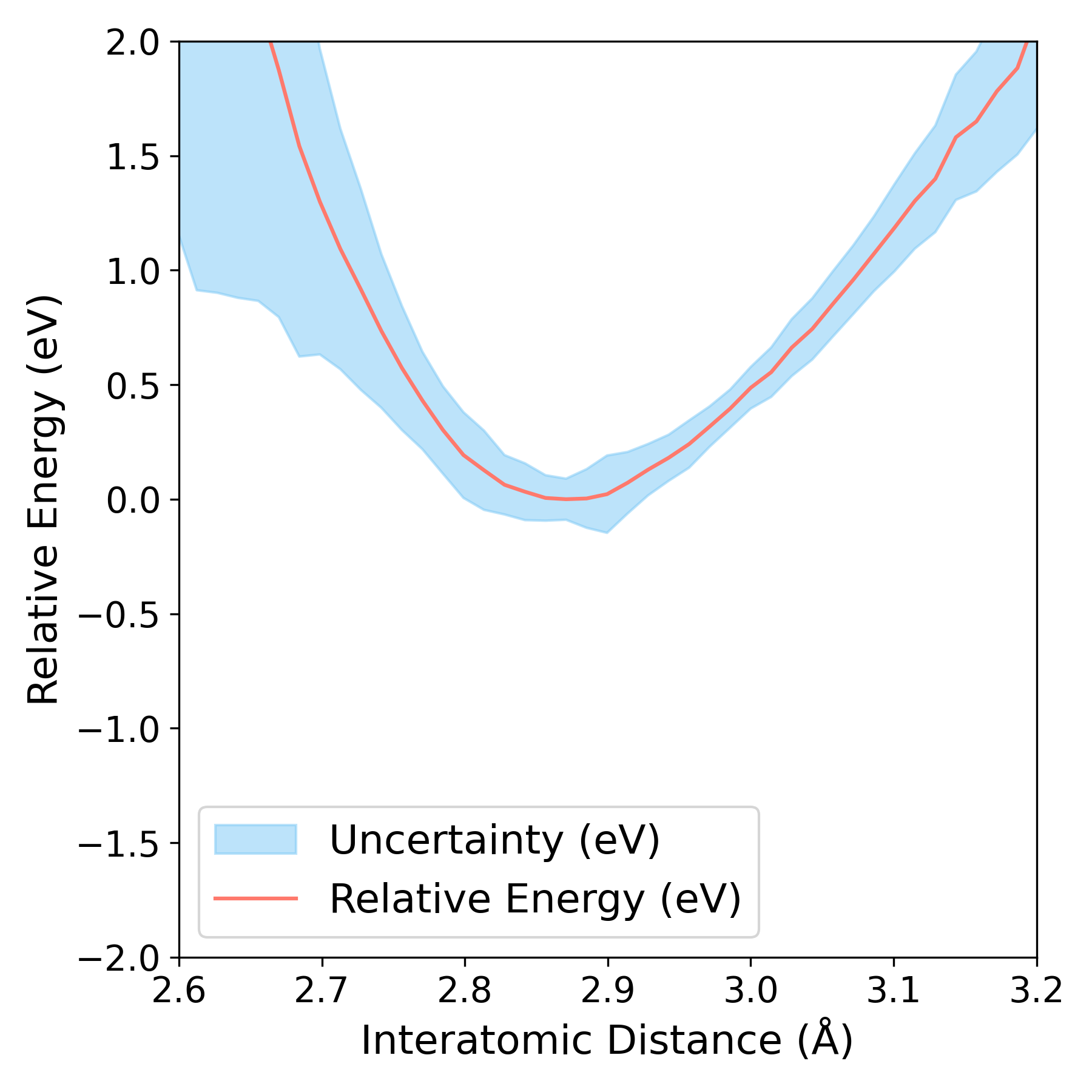}
     \caption{
        Al slab
     }
     \label{fig:eos_slab}
 \end{subfigure}
% ---------------end---------------
\caption{
Equation of state plots created using GemNet-OC, which describe the energy (relative to the plotted minimum) as a function of interatomic distance for three systems. \ref{fig:eos_cu} and \ref{fig:eos_tio2} are both bulk systems, which are not represented in the OC20 training data set. The predicted uncertainties for these systems are quite large, as expected. Note that the uncertainty scale is much larger \ref{fig:eos_tio2} which is an oxide bulk, and no oxide compositions of any kind are represented in the training set. \ref{fig:eos_slab} is an aluminum slab, which is similar to the slabs used to create the systems in the OC20 training data set, although this type of change to the interatomic distance of this slab is not represented in the training data. The predicted uncertainty for this system is significantly smaller, but grows towards the extremes, which also aligns with our expectations.
}
\label{fig:eos}
\end{figure}

For a more interpretable test case, we consider several specific examples of adsorbate and materials systems. 
The goal here is to check whether the latent space distance behaves as we would expect an uncertainty prediction method to behave. 
In Figure \ref{fig:eos} we look at how the predicted uncertainty (calibrated on relaxed adsorbate-slab energies) scales for predicting equation of states of two bulk materials, and an aluminum slab system. 
The predicted uncertainty is compared to the relative change in energy from the local minimum as the interatomic distance widens and contracts between the atoms in the metal structure. 
The underlying GemNet-OC model was trained exclusively on combinations of adsorbates and metal slabs present in the \gls{oc20} training dataset, and the latent distance uncertainty predictions were calibrated on similar systems. 
Since bulk materials, and oxide materials, are not present in the training data, we expect to see high uncertainty predictions for the Cu bulk example, and even higher for the Ti oxide bulk. 
This behavior is exactly what we observe: the fluctuations in energy due to changes in the lattice constant are completely dwarfed by the uncertainty predictions for both bulk systems.
We also expect the uncertainty predictions to be comparatively more certain about the slab system, which is what we observe.
The slab should be more similar than the bulk systems to the training data it has seen, despite this Al slab not being present in the training data.
We would expect the training data to contain many examples of stable slab structures, with near optimum interatomic distances.
Therefore we expect the \gls{gnn} to be fairly certain about its energy predictions near the stable interatomic distance for a given slab, relative to the rate of change in energy with respect to distance it predicts near the local minimum.
We also observe increased uncertainty further from the local minimum energy for interatomic distance, which is another promising sign.

\begin{figure}
\centering
\includegraphics[width=0.99\textwidth]{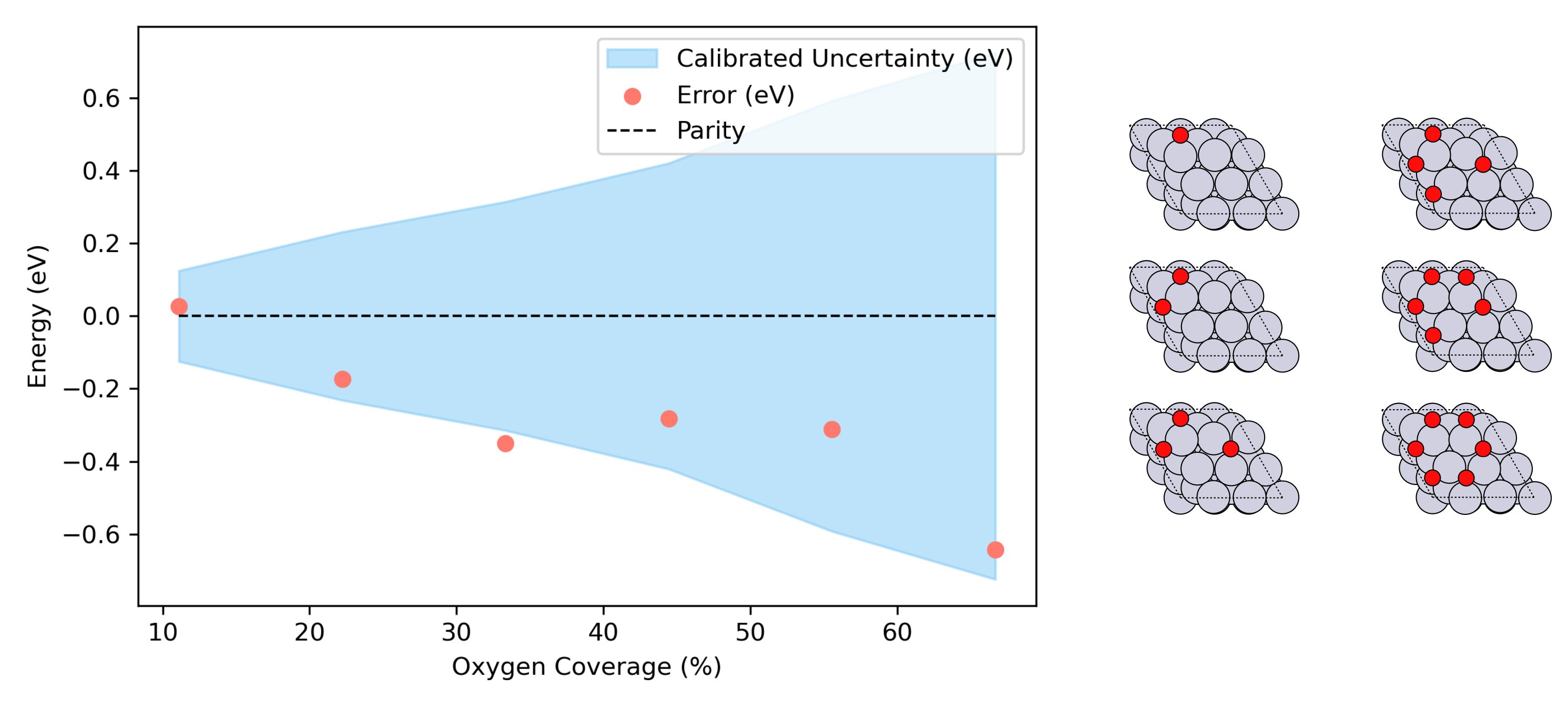}
\caption{
    Example using GemNet-OC with the latent distance uncertainty to predict platinum oxygen coverage. One oxygen adsorbate resting on the platinum slab should be very similar to examples seen in the training set, but with increasing coverage of oxygen atoms on the surface the system should appear to be further and further out of domain. The uncertainty boundary increases smoothly as expected as more oxygen adsorbates are added, and the error appears to increase in a similar manner.
}
\label{fig:pt_coverage}
\end{figure}

In Figure \ref{fig:pt_coverage} we look at another interpretable example, this time of oxygen coverage on a platinum surface.
Each system within the training and calibration data sets contains only a single adsorbate placement per unit cell, and therefore multiple adsorbate coverage on a single surface should appear to be increasingly out of domain for a useful uncertainty prediction.
The majority of unit cells in the training data set contain surfaces with multiple adsorption sites, but with adsorbate coverage on only one of them, so increasing the coverage on the surface should be novel from the perspective of the machine learning potential.
For this example we measure the actual error in the total energy predicted between\gls{vasp} and the machine learning potential, and compare it to the uncertainty prediction.
The error is measured after relaxation with the machine learning potential, since \gls{gnn} relaxed systems with adsorbates on slab surfaces is exactly what this \gls{rs2re} uncertainty method was calibrated for.
We expect to see the error and the uncertainty predictions increase commensurately with increasing coverage.
One key feature to note is that we expect the uncertainty prediction to increase smoothly and monotonically, which is exactly what we observe in Figure \ref{fig:pt_coverage}.
We also see that the magnitude of the errors does not increase perfectly monotonically, but does follow the general trend of increasing uncertainty, and this aligns with our expectation that the uncertainty predictions represent the dispersion of the distribution that the measured errors are sampled from.

\section{Conclusion}
Effective uncertainty prediction methods for GNN relaxed energies are key to the development of faster and more accurate screening techniques for novel material discovery. 
Quantifying the performance of uncertainty methods on relaxed energy predictions is especially complex, due to distribution assumptions built into most commonly employed UQ techniques.
Distribution-free techniques which employ bootstrapped confidence intervals, such as the CI(Var(Z)) test and error-based calibration plots, have been shown to be better metrics for analyzing the calibration of a UQ method in similar contexts, and we employ them here to great effect.
We show that latent distance methods outperform ensembles and other uncertainty methods on the \gls{rs2re} task, which is of practical relevance to workflows such as AdsorbML. 
We also show that the choice of latent representation is very important to the calibration of the latent distance as an uncertainty metric.
In the \gls{gnn} latent space, atom-wise distances produce better calibrated than system-wise distances.
Using rotationally invariant latent representations is crucial to producing calibrated distance measures, and the rotationally invariant latent space of GemNet-OC, a less accurate model, serves to compute a more well calibrated measure of uncertainty for EquiformerV2 than its own rotationally invariant latent space.
We compare the latent space uncertainty predictions of GemNet-OC to equation of state and coverage examples with expected outcomes, and find that its behavior aligns with our expectations of a useful uncertainty estimation method.
Finally, we challenge the community to improve on this \gls{rs2re} task for predicting uncertainties, using our proposed recalibration framework as a measure. 
Future work in this area should also explore the prediction of global minimum energy uncertainties directly, and the development of model architectures training methods or distance measures which preserve rotational equivariance while producing meaningful latent space distances. 

\section{Data and Software Availability}
    Details of the implementations, and additional tables of results, for the ensemble, distance, MVE, and sequence regression methods, as well as an examination of the error distributions, can be found in the supporting information.

\bibliography{reference}

\begin{tocentry}
\includegraphics[width=235.5pt]{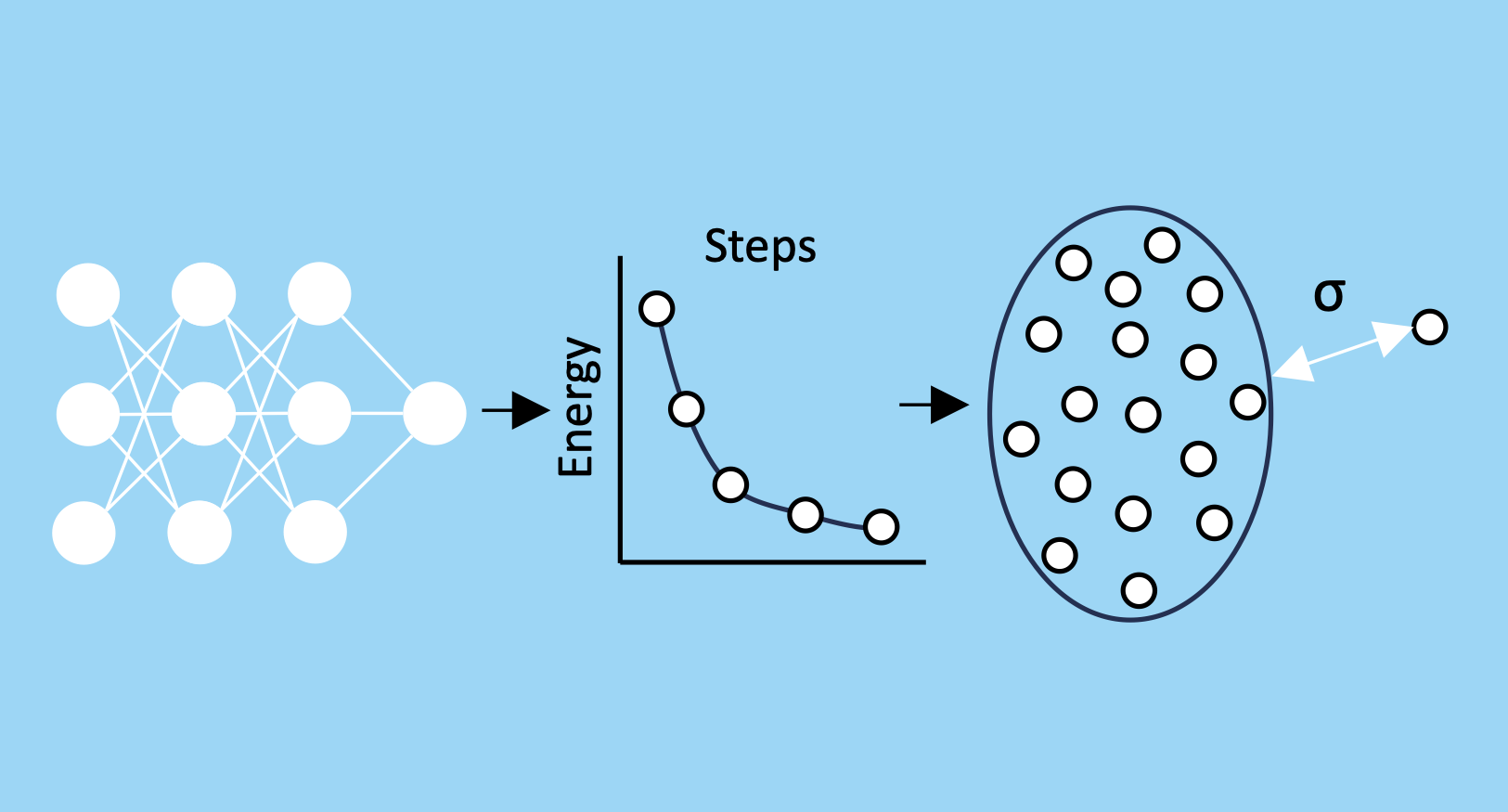}
TOC graphic
\end{tocentry}

\end{document}

\maketitle

\appendix

\section{Ensemble Methods}
\label{sec:ensembles_si}
The architecture ensemble contains eleven members, and uses several different \gls{gnn} architectures to achieve diversity, trained on largely the same data from the \gls{oc20} \gls{s2ef} training set. 
These model architectures include GemNet-OC, eSCN, and EquiformerV2.
The parameters ensemble contains six members, all using the EquiformerV2 architecture, trained on the entire \gls{oc20} training set, and varies only by the number of parameters used during training.
The bootstrap ensemble contains 10 identical members, using the EquiformerV2 architecture, each trained on a randomly selected slice (66\%) of the \gls{oc20} training set.
For each trajectory, we intend to predict a single uncertainty on the final frame, from the variance in the energy predictions of each of the members of the ensemble.
We assess the predicted uncertainty for the ensemble methods by computing the variance in the energy predictions across all members of the ensemble for each system.
We hypothesize that trajectory of all the energy predictions may contain some information relevant to predicting the uncertainty, so in addition to comparing across three ensembles, we also compare the effect of computing the uncertainty from the trajectory of variances.
We consider four methods, using the variance of only the first frame, only the last frame, the max variance over the whole trajectory, and the mean of the variances over the whole trajectory.
Taking the variances of all the frames over the whole trajectory contributes significantly more computational cost to this method.

\begin{table}[h]
\caption{Distribution free calibration metrics for each ensemble method tested. We construct three different ensembles, and test them using three different methods of predicting the relaxed energy uncertainty. Although some are close, none of the ensembles passed the CI(Var(Z)) test.}
\label{tab:ensembles}
\centering

\begin{tabular}{llrrrrl}
\toprule
 Ensemble     & Method   &   Parity $R^{2}$ &   Fit $R^{2}$ &   Slope &   Intercept & CI(Var(Z))   \\
\midrule
 Bootstrap    & First    &            0.809 &         0.896 &   1.056 &       0.065 & [1.52, 2.06] \\
 Bootstrap    & Last     &            0.838 &         0.914 &   0.897 &       0.104 & [1.63, 2.42] \\
 Bootstrap    & Max      &            0.817 &         0.890 &   1.025 &       0.074 & [1.61, 2.35] \\
 Bootstrap    & Mean     &   \textbf{0.870} &         0.930 &   0.912 &       0.094 & [1.58, 2.48] \\
 Parameters   & First    &            0.793 &         0.874 &   0.985 &       0.081 & [1.51, 1.99] \\
 Parameters   & Last     &            0.842 &         0.927 &   0.828 &       0.114 & [1.62, 2.37] \\
 Parameters   & Max      &            0.844 &         0.905 &   0.978 &       0.079 & [1.53, 2.19] \\
 Parameters   & Mean     &   \textbf{0.849} &         0.921 &   0.845 &       0.107 & [1.55, 2.33] \\
 Architecture & First    &            0.830 &         0.933 &   1.050 &       0.070 & [1.58, 2.19] \\
 Architecture & Last     &            0.891 &         0.942 &   0.945 &       0.081 & [1.55, 2.46] \\
 Architecture & Max      &            0.860 &         0.933 &   1.078 &       0.053 & [1.46, 2.29] \\
 Architecture & Mean     &   \textbf{0.905} &         0.940 &   0.946 &       0.070 & [1.42, 2.57] \\
\bottomrule
\end{tabular}

\end{table}

Table \ref{tab:ensembles} shows the choice of ensemble seems to be the most important to the success of the uncertainty prediction, followed by the method used to predict the uncertainty. For the better ensemble (architecture) all but the first-frame method beat every other ensemble-method combination in terms of parity $R^2$. This supports prior work which shows diversity between different members of an ensemble to be the most critical factor in producing good uncertainty predictions. Within each ensemble, there is a fairly consistent pattern of the mean method (taking the mean over the variances of all the frames in the trajectory) being the most effective method of computing uncertainty. This is followed by either the max method, or the last frame method, and then the first frame method is always the worst. This aligns with the notion that some additional information about the uncertainty of the model on a particular relaxed point can be gleaned from its uncertainty on other related points along the trajectory.

\section{Distance Methods}
\label{sec:distance_si}
The distance methods make use of the distance between points in some hidden latent space for each inference call. In every case, we extract the latent representation from some hidden layer after the interaction blocks in a \gls{gnn}. We then compute a latent space representation for every relaxed system in the training data set (460,000 systems), and relaxed systems from the in-domain and out-of-domain validation data sets (25,000 systems each). The in-domain validation data is used as a calibration set, while the out-of-domain validation data is used as the test set. For each data point in the calibration and test sets, we compute the shortest distance from that latent representation to any latent representation in the train set. These latent representations are calculated on a per-atom basis, with two options for computing distances. Either the distances between atom representations are computed, or the mean of the representations is computed and the distances between those means are computed. We make three different comparisons to assess the effects of different choices on the performance of latent space distance methods.

First we compare the effect of using the distance between the means of the latent representation over the entire system, to using the distances between the per-atom latent representations. For the per-atom latent representations, to reduce all the per-atom distances to a single value, we compare taking the mean of the distances, the sum of the distances, or the maximum of the distances. Across the different versions of the EquiformerV2 latent space that we sampled, we found that any version of the per-atom latent distances universally outperformed the per-system latent distance. Within the per-atom latent distance approaches, we found the mean of the per-atoms distances to be often the best performer, and reliably high-performing across all models and latent representations. These results can be found in the main text.

Second we compare the effect of different latent space sampling methods for expressing rotational equivariance and/or invariance in the latent representation. We do this by sampling latent representations from EquiformerV2 from certain spherical channels, and before and after edge alignment. We take three approaches to sampling spherical channels: sampling only the l=0, m=0 channel, sampling all the m=0 channels, and sampling all the channels. The l=0, m=0 channel should be inherently invariant to rotation. All of the m=0 channels should be sensitive to some rotations when the representation is subject to random rotations, but during the output blocks, the representation is reliably rotated to be aligned with the edges of the input graph, and therefore these channels should behave as though they are invariant to rotation. Finally, the representation of all of the channels should always be equivariant to rotation \cite{equiformerv2}. We find that using all the channels for the latent space is consistently worse than using only the invariant channels. And that using the edge-aligned version of the l=0, m=0 performs best. These results can be found in Table \ref{tab:distance_eq2_methods}.

\begin{table}[h]
\caption{Comparison of latent space distances extracted from Equiformer V2 using different approaches aimed at eliminating rotational equivariance, all used to predict the error of Equiformer V2 on relaxed structures.}
\label{tab:distance_eq2_methods}
\centering

\begin{tabular}{llrrrrl}
\toprule
 Model        & Sphharms          &   Parity $R^{2}$ &   Fit $R^{2}$ &   Slope &   Intercept & CI(Var(Z))   \\
\midrule
 EqV2 (nodes) & m=2, l=4 (equiv.) &            0.813 &         0.902 &   0.974 &       0.060 & [1.51, 5.45] \\
 EqV2 (edges) & m=2, l=4 (equiv.) &            0.745 &         0.957 &   1.211 &       0.008 & [1.47, 2.45] \\
 EqV2 (nodes) & m=0, l=0 (inv.)   &            0.866 &         0.913 &   0.974 &      -0.037 & [1.04, 5.44] \\
 EqV2 (edges) & m=0, l=0 (inv.)   &   \textbf{0.910} &         0.911 &   1.014 &      -0.002 & [1.34, 6.37] \\
 EqV2 (edges) & m=0, l=4          &            0.906 &         0.913 &   0.998 &      -0.020 & [1.10, 5.16] \\
\bottomrule
\end{tabular}

\end{table}

Third, we test 4 different \gls{gnn}s for their ability to predict the uncertainty of \gls{rs2re} predictions made using EquiformerV2: PAINN, eSCN, EquiformerV2, and GNOC \cite{escn, painn, gemnetoc, equiformerv2}. 
PAINN, eSCN, and EquiformerV2 all take a rotationally equivariant approach, while GNOC preserves rotational invariance throughout the model. 
In each case, we sample the entire latent space of the \gls{gnn} immediately after the final interaction block.
We find that GNOC outperforms even the invariant latent space of EquiformerV2 at predicting EquiformerV2's uncertainty.
These results can be found in Table \ref{tab:distance_models}.

\begin{table}[h]
\caption{Comparison of latent space distances extracted from different models, all used to predict the error of Equiformer V2 on relaxed structures.}
\label{tab:distance_models}
\centering

\begin{tabular}{llrrrrl}
\toprule
 Model       & Latent Rep.   &   Parity $R^{2}$ &   Fit $R^{2}$ &   Slope &   Intercept & CI(Var(Z))    \\
\midrule
 EqV2 equiv. & nodes         &            0.813 &         0.902 &   0.974 &       0.060 & [1.51, 5.45]  \\
 EqV2 inv.   & nodes         &            0.866 &         0.913 &   0.974 &      -0.037 & [1.04, 5.44]  \\
 EqV2 inv.   & edges         &            0.910 &         0.911 &   1.014 &      -0.002 & [1.34, 6.37]  \\
 eSCN        & nodes         &            0.818 &         0.931 &   1.090 &       0.021 & [1.38, 4.90]  \\
 PAINN       & nodes         &            0.856 &         0.934 &   0.971 &      -0.050 & [0.72, 1.21]* \\
 GNOC        & nodes         &   \textbf{0.932} &         0.965 &   1.108 &       0.006 & [1.17, 2.18]  \\
\bottomrule
\end{tabular}

\end{table}

\section{MVE and Sequence Regression Methods}
\label{sec:mve_si}
The MVE and sequence regression methods we tested both aim to directly predict the uncertainty of EquiformerV2 by training a neural network, or portion of a neural network.
These models take the full latent representation at the end of the last interaction block, and are fit on  residuals of the EquiformerV2's energy predictions.
In the case of the MVE methods, an additional output head, or an ensemble of output heads is added to the fully trained EquiformerV2 checkpoint.
These new output heads are initialized randomly, the rest of the model is frozen, and they are trained on the residuals of the in-domain validation data, until the performance of the direct residual prediction stops improving on a held out portion of the validation data set.
In the case of the single head, the loss is computed as a the difference between its direct prediction and the residual values.
In the case of the ensemble of heads, the loss is computed as the difference between the variance in each of the ten heads energy predictions, and the residual values.

The sequence regression models are similarly trained using the full latent representations as input.
However, we hypothesize that change in the latent representation over the entire trajectory might contain information relevant to the task of directly learning the residuals on the last frame.
We use a transformer sequence regression model, as implemented in Hugging Face \cite{distilbert}.
We modify the transformer to accept vectors of latent representations for each atom as input, and batch over all of the atoms in the system.
Then for each trajectory, we train it on the sequence of atoms, batching over all the atoms, and regress to fit the residual on the final frame of the EquiformerV2 trajectory.
Similar to the MVE approach, we train on the in-domain validation data set, until performance stops improving on a held out portion of the data set.
As in all other cases, we recalibrate all uncertainty predictions on the out-of-domain validation data set, using error-based recalibration.
We compare the best performing result for each of these implementations in Table \ref{tab:uq_top_per_metric}.
However we note that both of these direct residual fitting methods had a tendency to overfit, performing much more poorly on the out-of-domain validation data than on the in-domain validation data, despite using a held-out portion of the validation data to perform early stopping.

\begin{table}
\caption{Best performers for each of the \gls{uq} validation metrics. Many of the popular \gls{uq} validation metrics assume normally distributed errors, such as miscalibration area ($A_{mis}$), before and after error-based recalibration, and the negative log likelihood (NLL). Other metrics target properties other than calibration, such as \spearman ($\rho$) and the area under the receiver operating characteristic curve (AUROC). These methods tend to be inconsistent with one another. The error based calibration (Parity $R^{2}$) measure and the CI(Var(Z)) test both measure calibration without a distribution assumption, and agree that the GNOC distance method performs best.}
\label{tab:uq_top_per_metric}
\centering

\begin{tabular}{lrrrrrrl}
\toprule
 Method            &   Parity $R^{2}$ &   $A_{mis,u}$ &   $A_{mis,r}$ &   $NLL$ &   $\rho$ &   AUROC & CI(Var(Z))   \\
\midrule
 GNOC distance     &            \textbf{0.952} &         0.365 &         0.178 &   0.279 &    0.468 &   0.755 & \textbf{[1.13, 1.79]} \\
 Architecture ens. &            0.830 &         \textbf{0.050} &         0.156 &   0.411 &    0.453 &   0.746 & [1.58, 2.19] \\
 Residual model    &            0.331 &         0.174 &         \textbf{0.102} &   1.170 &    0.360 &   0.695 & [3.38, 4.47] \\
 MVE head          &            0.831 &         0.340 &         0.114 &   \textbf{0.251} &    0.526 &   0.785 & [1.59, 2.40] \\
 Bootstrap ens.    &            0.870 &         0.248 &         0.121 &   0.303 &    \textbf{0.585} &   \textbf{0.817} & [1.58, 2.48] \\
\bottomrule
\end{tabular}

\end{table}

\section{Error Distribution}

\begin{figure}

 \begin{subfigure}{0.49\textwidth}
     \includegraphics[width=\textwidth]{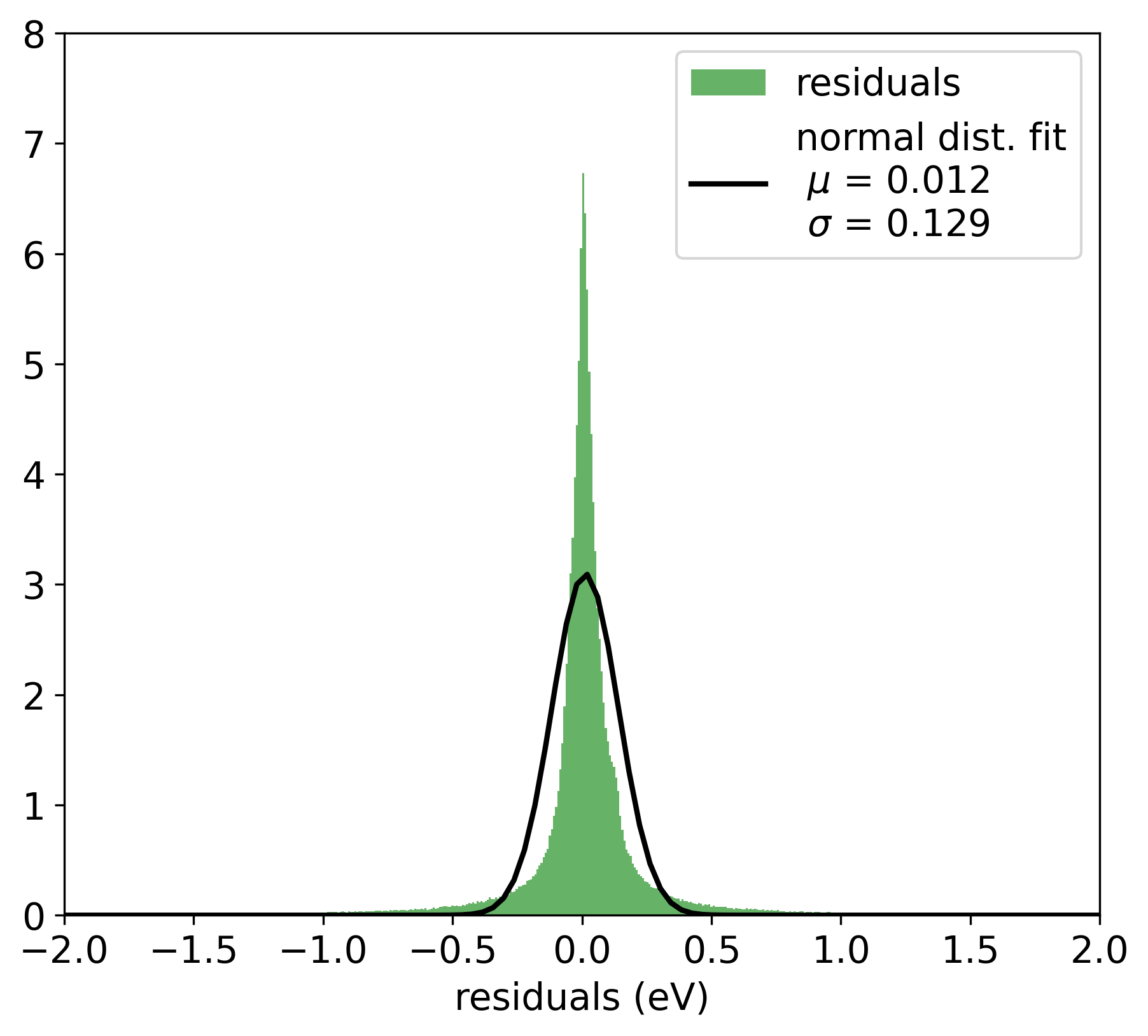}
     \caption{RS2RE error distribution}
     \label{fig:rs2re_distribution}
 \end{subfigure}
 \hfill
 \begin{subfigure}{0.49\textwidth}
     \includegraphics[width=\textwidth]{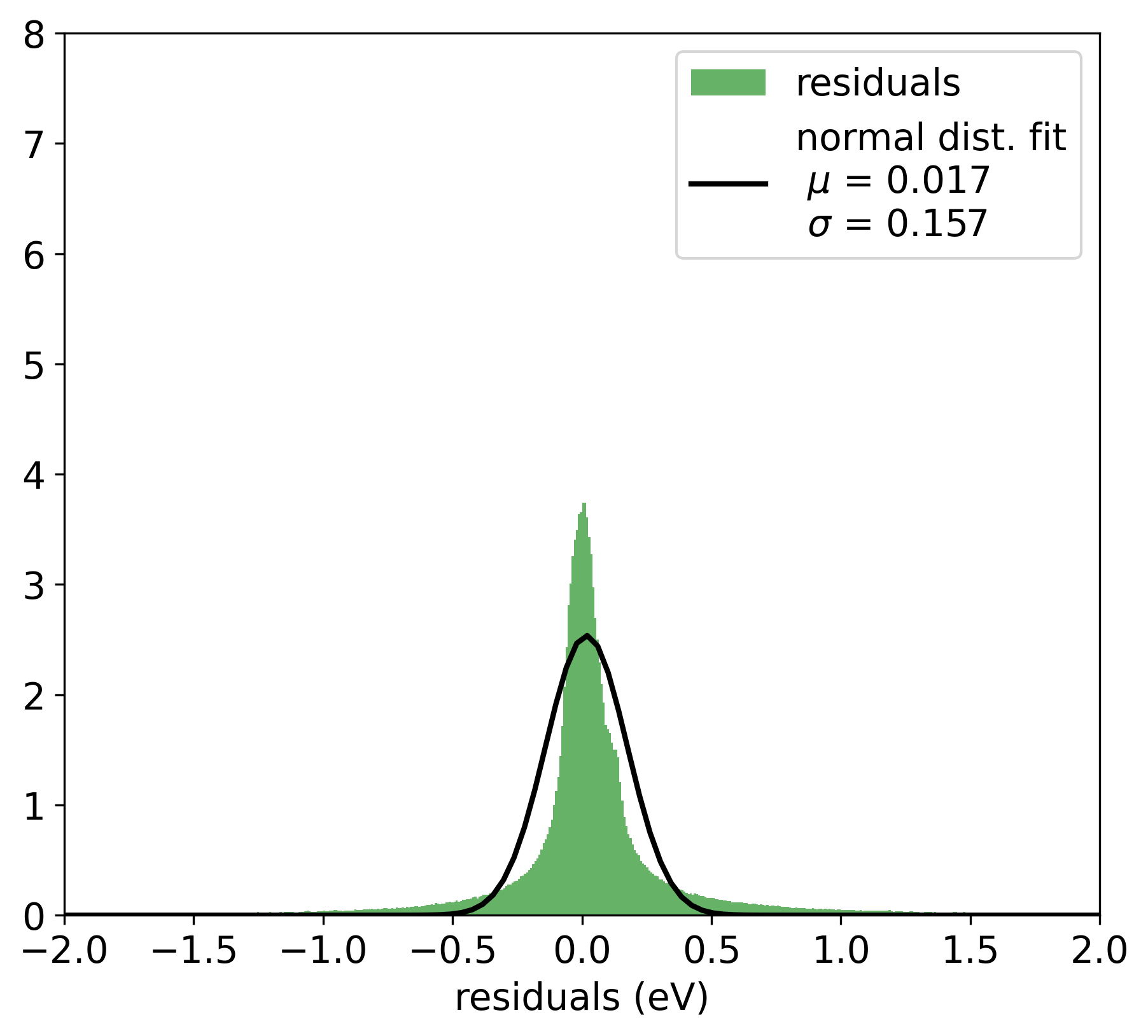}
     \caption{S2EF error distribution}
     \label{fig:s2ef_distribution}
 \end{subfigure}

 \caption{Density histograms of the S2EF and RS2RE error distributions. Both use the Equiformer V2 31M model checkpoint to make energy predictions, and the errors are the difference between the predicted energy and the \gls{vasp} ground truth energies.}
 \label{fig:histograms}

\end{figure}

We compare the \gls{s2ef} task, which involves predicting the energies and forces of points sampled indiscriminately from many \gls{dft} relaxations, to the \gls{rs2re} task, where energies are predicted only on points relaxed by a machine learning potential.
We plot the distribution of errors for both the \gls{s2ef} task and the \gls{rs2re} in Figure \ref{fig:histograms}, where we see the \gls{rs2re} error distribution is much more concentrated at 0 eV, and much further from a Gaussian distribution than the \gls{s2ef} task for the same model.
Predicting the uncertainty of a machine learning potential on the broader \gls{s2ef} task is a less challenging task than uncertainty prediction for \gls{rs2re}.
We believe this is the result of selecting points only from the end of the structural optimization process creating a non-Gaussian distribution of errors from the predictions.
This can be seen in Table \ref{tab:s2ef_rs2re}, where each of the commonly used \gls{uq} validation metrics agree that the same uncertainty quantification methods show poorer performance on the \gls{rs2re} task.

\begin{table}[htbp]
\caption{Comparison of ensemble uncertainty metrics for S2EF and RS2RE tasks. The S2EF task is consistently easier for the ensemble to perform well at predicting the uncertainty.}
\label{tab:s2ef_rs2re}
\centering

\begin{tabular}{llrrrrl}
\toprule
 Method       & Task   &   Parity $R^{2}$ &   $NLL$ &   $\rho$ &   $A_{mis}$ & CI(Var(Z))   \\
\midrule
 bootstrap    & RS2RE  &            0.822 &   0.366 &    0.569 &       0.144 & [1.74, 2.56] \\
 bootstrap    & S2EF   &            0.906 &   0.055 &    0.673 &       0.051 & [1.48, 1.52] \\
 architecture & RS2RE  &            0.869 &   0.380 &    0.604 &       0.136 & [1.62, 2.66] \\
 architecture & S2EF   &            0.958 &   0.075 &    0.670 &       0.023 & [1.32, 1.35] \\
\bottomrule
\end{tabular}

\end{table}

\clearpage
\bibliography{reference}